%% file: main.tex
\PassOptionsToPackage{table}{xcolor}
\documentclass[10pt,twocolumn,letterpaper]{article}

\usepackage[pagenumbers]{iccv} % To force page numbers, e.g. for an arXiv version

\input{preamble}

\definecolor{iccvblue}{rgb}{0.21,0.49,0.74}
\usepackage[pagebackref,breaklinks,colorlinks,allcolors=iccvblue]{hyperref}

\def\paperID{1243} % *** Enter the Paper ID here
\def\confName{ICCV}
\def\confYear{2025}

\title{Event-Driven Storytelling with Multiple Lifelike Humans in a 3D Scene}

\author{
    Donggeun Lim$^1$ \quad Jinseok Bae$^1$ \quad Inwoo Hwang$^1$ \quad Seungmin Lee$^1$\\
    Hwanhee Lee$^2$ \quad Young Min Kim$^1$\\[0.2em]
    \normalsize{$^1$Seoul National University \quad $^2$Chung-Ang University}\\
    \texttt{\small $^1$\{rms2836, capoo95, inusu0818, rsual, youngmin.kim\}@snu.ac.kr} \quad \texttt{\small $^2$\{hwanheelee\}@cau.ac.kr}
}

\begin{document}
% \maketitle

% (https://github.com/cvpr-org/author-kit/issues/25#issuecomment-1815782690)
\twocolumn[{%
\renewcommand\twocolumn[1][]{#1}%
\maketitle
\begin{center}
    \centering
    \captionsetup{type=figure}
    \includegraphics[width=1.0\linewidth]{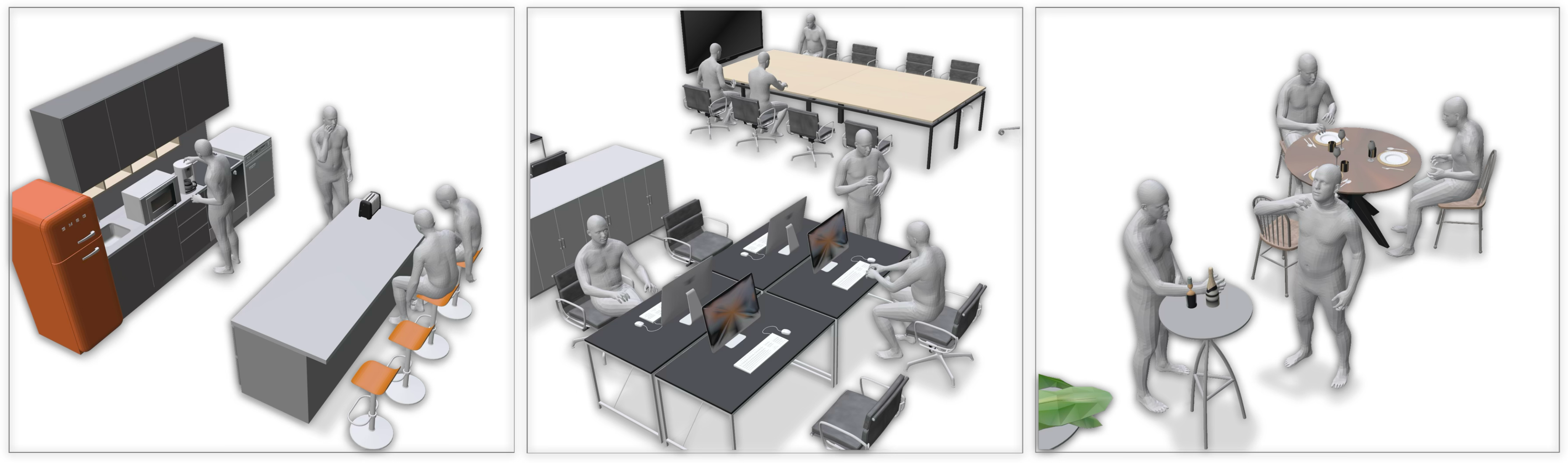}
    % \vspace{-1em}
    \captionsetup{width=\linewidth}
    \captionof{figure}{Our framework populates 3D scenes with multiple characters. The generated characters interact with their surroundings and other people, bringing the space to life.}
    \label{fig:teaser}
\end{center}%
}]

\input{sec/0_abstract}    
\input{sec/1_intro}
\input{sec/2_related}
\input{sec/3_method}

\input{sec/4_experiment}
\input{sec/5_conclusion}
\input{sec/6_supp}

{
    \small
    \bibliographystyle{ieeenat_fullname}
    \bibliography{main}
}

\end{document}

%% file: preamble.tex
\usepackage{xcolor}
\usepackage{listings}
\usepackage{multirow}
\usepackage{booktabs}
\usepackage{algorithm}
\usepackage{algpseudocode}

\definecolor{kewordcolor}{rgb}{0.156,0.411,0.513}
\definecolor{functioncolor}{rgb}{0.337,0.580,0.623}
\definecolor{commentcolor}{rgb}{0.42,0.42,0.396}
\lstdefinestyle{customPython}{
    language=Python,
    backgroundcolor=\color{white},
    basicstyle=\ttfamily\scriptsize,
    keywordstyle=\bfseries\color{kewordcolor},
    stringstyle=\color{black},
    commentstyle=\color{commentcolor},
    breaklines=true,
    frame=single,
    rulecolor=\color{gray},
    morekeywords={[3]get_area_to_sit_on,get_character,set_position,set_target_action,parse_event},
    keywordstyle=[3]\color{functioncolor},
    morekeywords={[2]get_distance_between},
    keywordstyle=[2]\bfseries\color{functioncolor},
}

%% file: sec/0_abstract.tex
\begin{abstract}

In this work, we propose a framework that creates a lively virtual dynamic scene with contextual motions of multiple humans.
Generating multi-human contextual motion requires holistic reasoning over dynamic relationships among human-human and human-scene interactions.
We adapt the power of a large language model (LLM) to digest the contextual complexity within textual input and convert the task into tangible subproblems such that we can generate multi-agent behavior beyond the scale that was not considered before.
Specifically, our event generator formulates the temporal progression of a dynamic scene into a sequence of small events.
Each event calls for a well-defined motion involving relevant characters and objects.
Next, we synthesize the motions of characters at positions sampled based on spatial guidance.
We employ a high-level module to deliver scalable yet comprehensive context, translating events into relative descriptions that enable the retrieval of precise coordinates.
As the first to address this problem at scale and with diversity, we offer a benchmark to assess diverse aspects of contextual reasoning.
Benchmark results and user studies show that our framework effectively captures scene context with high scalability.
The code and benchmark, along with result videos, are available at our \href{https://rms0329.github.io/Event-Driven-Storytelling/}{project page}.

\end{abstract}

%% file: sec/1_intro.tex
\section{Introduction}
\label{sec:intro}
A scene with digital humans creates lively atmosphere and has a wide range of applications in VR, games, and movies.
The realism incurs from natural motions of multiple characters that seamlessly integrate into the surrounding environment.
However, existing motion synthesis works focus on individual characters independently or extend to interactions with either the surrounding environment~\cite{hassan2021stochastic, zhao2023synthesizing, mir2024generating, wang2021synthesizing, wang2022humanise, wang2022towards, araujo2023circle, wang2024move, yi2024generating, jiang2024autonomous, cen2024generating} or another human~\cite{liang2024intergen, ghosh2024remos, shan2024towards, tanaka2023role, liu2024physreaction}, remaining limited to a single type of interaction.
Hence, they severely lack generalizability and scalability to concurrently generate plausible interactions of multiple characters in a crowded scene.
It is highly challenging to correctly assign the behavior of individual characters at the correct time stamp as it requires extensive dynamic contextual and spatial reasoning.

To fulfill such holistic requirements, we take inspiration from recent advances in large language models (LLMs).
LLMs demonstrated phenomenal performance in various areas, especially for high-level planning~\cite{liang2023code, singh2023progprompt, wangvoyager, huang2022language, rana2023sayplan, gramopadhye2023generating} or behavior modeling~\cite{park2023generative, cai2024digital}. 
We deduce that LLMs have the potential to interpret nuanced semantics from open-ended textual inputs and dependency between natural motion and scene context.
However, simply generating commands from the input request may suffer from hallucinations or struggle with localization errors, which are well-known limitations of foundation models~\cite{xu2024hallucination, mirzadeh2024gsm, shi2023large, jiang2024peek}.

We propose an LLM-powered framework that populates multiple lifelike virtual humans within a 3D scene.
As shown in Figure~\ref{fig:teaser}, the characters produced by our framework can understand and interact with their surroundings, including both 3D scenes and other characters.
The proposed LLM module can flexibly understand the holistic context in open-ended scenarios and make high-level decisions.
The characters may exhibit emergent yet plausible behavior that was not explicitly defined within the original input, providing a lively atmosphere with rich interactions.
The framework can further adapt to user-interactive scenarios, where users may interact with the characters through high-level text instructions.

Towards more performant and scalable planning with LLMs, our modular framework is deliberately designed to reduce the reasoning complexity.
We introduce an event-based planning approach with two LLM modules: \textit{narrator} and \textit{event parser}.
The \textit{narrator} progressively weaves the flow of the scene by generating a textual description of one event at a time. 
Each event considers only a well-defined subset of characters and objects, detached from the collective behavior of the holistic flow.
The \textit{event parser} then transforms high-level event representation into detailed information for existing motion synthesis frameworks.

We provide devised scene information tailored for individual modules to guide necessary spatial reasoning.
Our \textit{scene describer} converts the 3D scene graph into a textual summary and provides it as input to the narrator and the event parser.
Our prompt encourages contextual information on different utilization of spaces, such that the description delivers the overall structure.
The event parser provides accurate 3D scene grounding to the low-level motion synthesis module, such that it can generate fine-grained and realistic interaction.
Inspired by previous studies~\cite{liang2023code, yang2024llm, wangvoyager, singh2023progprompt}, the event parser employs spatial reasoning tools for LLM and specifies a relevant location within a programming framework.
Furthermore, we propose an area-conditioned position sampling technique to compensate for the weakness of LLMs in precise localization at the coordinate level.

Our framework successfully generates long-term motions with more than 4-5 characters in various multi-room scale scenes, and is robust to the choice of LLM engines.
We also propose evaluation criteria for scene-aware multi-agent behavior planning tasks and create a benchmark to assess the compatibility of our framework comprehensively.
The extensive results demonstrate that our proposed pipeline outperforms in generating a natural and plausible sequence of actions with adequate trajectories at appropriate relative timing in a shared space, especially for larger scenes with more number of characters.

%% file: sec/2_related.tex
\section{Related Works}
\label{sec:related}

\paragraph{Contextual Human Motion Synthesis}
Synthesizing human motions is a long-standing research topic in the fields of computer vision and graphics. 
In particular, an increasing number of studies have incorporated contextual information in motion synthesis, such as a 3D environment~\cite{hassan2021stochastic, zhao2023synthesizing, mir2024generating, wang2021synthesizing, wang2022humanise, wang2022towards, araujo2023circle, wang2024move, yi2024generating, jiang2024autonomous, cen2024generating} or another human~\cite{liang2024intergen, ghosh2024remos, shan2024towards, tanaka2023role, liu2024physreaction}.
Although these studies have shown notable progress in reproducing physically accurate and high-fidelity motions for a single type of interaction, most of them do not consider more than one contextual constraint or expand to long-term sequences of actions.
Some recent works~\cite{cai2024digital, chen2024sitcom} have explored motion generation considering both humans and scenes of contextual information.
\textit{Digital Life Project}~\cite{cai2024digital} builds autonomous characters with social intelligence in a 3D scene, while \textit{Sitcom-Crafter}~\cite{chen2024sitcom} introduces a unified motion generation framework that integrates different types of interaction motions with plot-driven guidance.
The research expanded on previous boundaries, yet it did not take into account the coordination of interactions between more than two characters.
Furthermore, in terms of scene understanding, these models lack the complexity needed for scenarios requiring detailed spatial reasoning because of their simplistic scene descriptions.
Compared to these studies, our framework demonstrates successful multi-agent contextual behavior planning at a larger scale, including an increased number of characters and a scene with greater size and complexity.

\paragraph{LLM-based Planning}
In recent years, there has been growing attention to using pre-trained LLMs in various domains for their impressive zero-shot reasoning capabilities~\cite{hurst2024gpt, dubey2024llama, yang2024qwen2}.
Particularly in the field of robotics and embodied AI, many recent works have successfully demonstrated the abilities of LLMs for task planning~\cite{gramopadhye2023generating, liang2023code, wangvoyager, singh2023progprompt, ni2024grid, huang2022language, li2023interactive, rana2023sayplan, kannan2024smart}.
However, robotic applications primarily focus on goal-oriented tasks, which do not require complex contextual reasoning.
While the majority of these works are limited to single-agent scenarios, a few studies~\cite{zhangbuilding, kannan2024smart} have explored LLM-based multi-agent task planning recently.
\textit{CoELA}~\cite{zhangbuilding} introduces a novel cognitive-inspired modular framework that enables cooperation and communication between multiple agents.
However, since \textit{CoELA} requires continuous message exchanges between agents for behavior coordination, this can result in inefficiency when scaling to an increased number of agents.
On the other hand, \textit{SMART-LLM}~\cite{kannan2024smart} demonstrated multi-agent task planning in larger-scale scenarios.
\textit{SMART-LLM} accomplishes multi-robot cooperation by decomposing tasks into several sub-stages.
However, \textit{SMART-LLM} represents a scene as a mere list of objects, leading to a lack of comprehensive scene understanding.
Distinct from earlier approaches, ours tackles the problem of multi-agent planning while maintaining scalability and in-depth scene understanding.

%% file: sec/3_method.tex
\begin{figure}[t]
\centering
\includegraphics[width=1.05\linewidth]{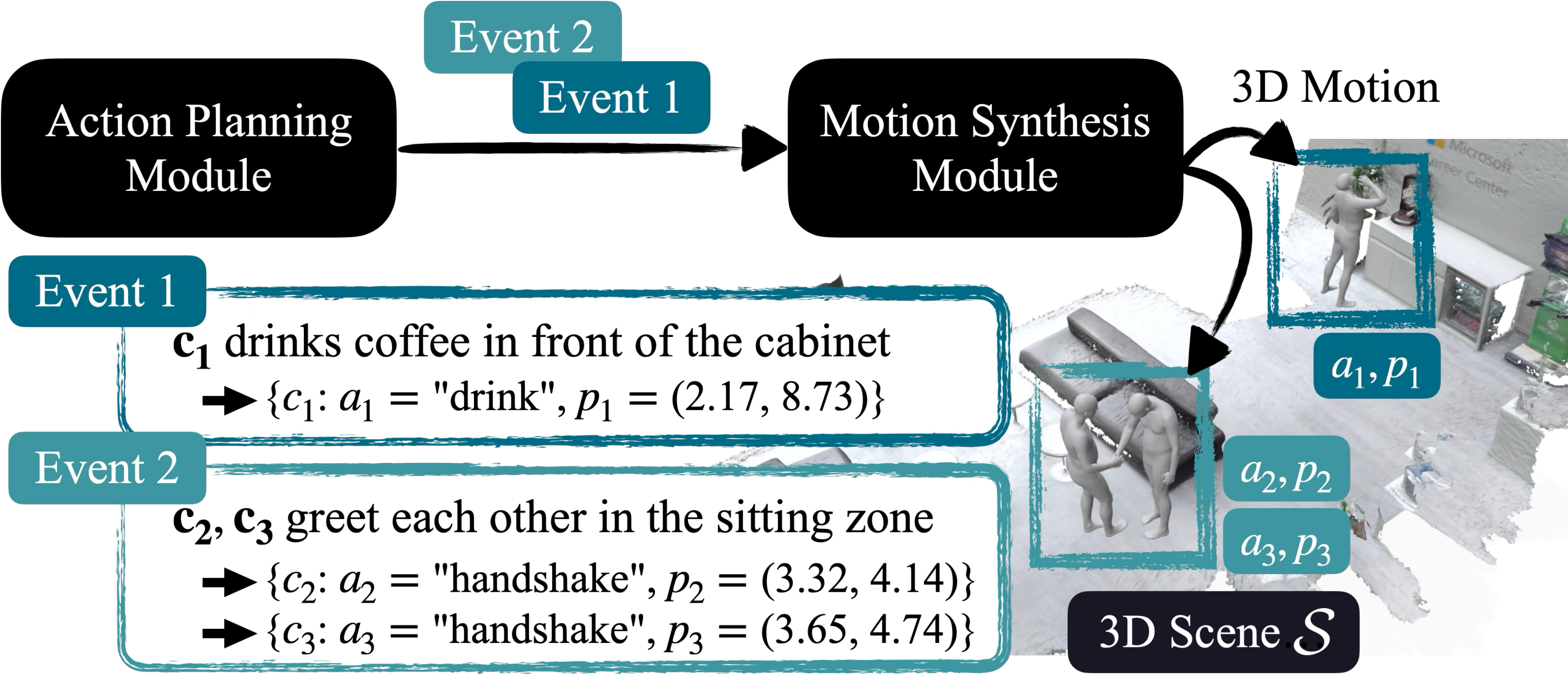}
\caption{Overview of our framework. The framework consists of two main modules: a high-level action planning module and a low-level motion synthesis module. 
The action planning module coordinates and plans multi-character behavior as a sequence of events.
The motion synthesis module receives a detailed description of the event and synthesizes 3D motions.
}
\label{fig:overview}
\vspace{-1.5em}
\end{figure}

\section{Method}
\label{sec:method}
Given a 3D scene $\mathcal{S}$, characters $\mathcal{C}=\{c_1, c_2, \cdots, c_N\}$, and optional user instructions $\mathcal{T}$, our framework generates 3D multi-human motions $\mathcal{M}=\{M_1, M_2,\cdots, M_N\}$ within the scene.
Here, $N$ represents the number of characters and is determined by the user at the system initialization.
The 3D scene $\mathcal{S}$ is provided with instance segmentation, and can be in various representations including VR scenes with CAD models or real-world scenes composed of 3D point cloud scans or mesh reconstructions.
The user instructions $\mathcal{T}$ are provided in free text forms, leveraging the generalizability of LLM engines.

In an overview, our framework operates around an intermediate representation called \textit{event} $e$.
Figure~\ref{fig:overview} contains examples of textual descriptions of events along with detailed action labels and positions for the motion synthesis module.
The events are generated in sequence to realize the given instruction while coordinating with the surrounding environments.
Each event has its own lifecycle with a different time span, and multiple events can coexist within the 3D scene.
Through the event-based formulation, our system can generate scene dynamics with infinitely long-horizon, and adaptively process user instructions at an arbitrary time during operation.

As shown in Figure~\ref{fig:overview}, our framework consists of two main components: a high-level action planning module and a low-level motion synthesis module.
The action planning module performs planning on an event basis, driving the scene's timeline forward (Section~\ref{sec:action_planning_module}).
The motion synthesis module takes the generated events as input, assigns them to the corresponding characters, and creates motions that reflect the assigned event (Section~\ref{sec:motion_synthesis_module}).

\begin{figure*}[t]
  \centering
   \includegraphics[width=0.95\linewidth]{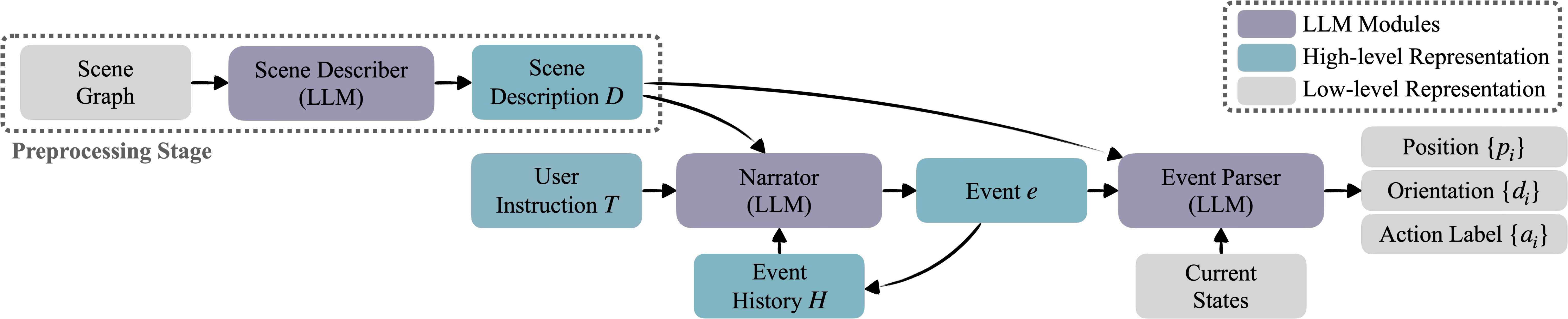}
   \caption{Overview of the high-level action planning module.}
   \label{fig:action_planning}
\vspace{-1em}
\end{figure*}

\subsection{High-level Action Planning Module}
\label{sec:action_planning_module}
The primary function of the action planning module is to generate new events by comprehensively considering the given 3D scene $\mathcal{S}$, existing event history $\mathcal{H}$, and the behaviors of multiple characters within the scene.
To effectively satisfy the holistic requirements, we rely on the powerful reasoning capabilities of LLMs.
As shown in Figure~\ref{fig:action_planning}, there are three LLM submodules: the scene describer, the narrator, and the event parser.
The scene describer generates a textual scene description $\mathcal{D}$ from the given 3D scene $\mathcal{S}$ such that our planning modules can understand the necessary context from the spatial arrangement.
The narrator generates a sequence of events, which are converted into detailed information by the event parser.
Each module is assigned to smaller, well-defined tasks such that the system stays performant and scalable for multi-human motion generation.

\subsubsection{Scene Describer}
\label{sec:scene_describer}
In the preprocessing stage before the motion planning, the scene describer generates a textual description $\mathcal{D}$ of the given 3D scene $\mathcal{S}$.
We use a 3D scene graph, which may lead an LLM module to absorb rich spatial relationships~\cite {huang2024embodied}.
The scene describer first extracts a 3D scene graph in an automated way as proposed in~\cite{jia2024sceneverse} and converts it into textual data in JSON format.
Then, we use it to prompt the LLM module and generate a scene description.
In addition to the explicit pairwise relationships in the scene graph, further contextual layouts with regional information enhance the naturalness of scene-aware motions.
To enhance contextual reasoning, we encourage our scene describer to discover a spatial arrangement of objects in proximity, which semantically creates a functional space, for example, a dining area or study zone.
Specifically, we extract a regional cluster of objects through the DBSCAN algorithm~\cite{ester1996density} as an additional input and promote detecting areas of interest.
We also provide concrete examples of the same input and output format, allowing the scene describer to perform the given task more effectively through in-context learning~\cite{brown2020language}.
We provide additional details and full prompts in the supplementary material.

\subsubsection{Narrator}
As the core of our behavior planning, the narrator generates a sequence of events $\mathcal{E}=\{e_1, e_2, ...\}$.
Given the scene description $\mathcal{D}$ from the scene describer and the optional user instruction $\mathcal{T}$, the narrator keeps the previous event history $\mathcal{H}$ and generates one event at a time based on the high-level context.
Focusing on one event at a time, the subsequent module can alleviate the burden of holistic spatio-temporal reasoning, making the storytelling scalable and interactive.
The event-based design also allows the incorporation of additional user instructions between events to refine the storyline.

As shown in Figure~\ref{fig:overview}, events are expressed in semi-narrative natural language, focusing on contextual description.
The textual description is further parsed in the subsequent event parser, relieving the burden of detailed spatio-temporal localization.
The narrator also receives the state of a generated event as one of two states, `ongoing' or `completed'.
From this, the narrator can compactly conceive the necessary information to progress the timeline.
Our event representation is particularly efficient for group events with multiple characters, which makes the broad scene context easily understood.
The narrator coordinates the high-level behaviors of multiple characters and generates the storyline from a macroscopic perspective.
The detailed spatio-temporal progression of individual characters is assigned to the event parser.

Similar to the scene describer, we also incorporate several widely used prompting techniques in the narrator's prompt, the in-context learning and chain-of-thought reasoning~\cite{wei2022chain}.
% We incorporate in-context learning and chain-of-thought reasoning~\cite{wei2022chain} techniques in the narrator's prompt.
For chain-of-thought reasoning, we guide the narrator to first reason about the current planning state and other characters’ statuses before generating an event.
We provide concrete examples of event generation with such a reasoning process to reinforce structured decision-making and improve the coherence of generated events.
Further details and the full prompts are provided in the supplementary material.

\begin{figure}[t]
    \centering
    \begin{minipage}{1.0\linewidth}    
    \begin{lstlisting}[style=customPython]
"""
Event to parse:
[Sara] drinks a beverage while sitting in the seat farthest from the reception desk
"""
def parse_event():
    desk = "reception_desk_1"
    chiars = ["chair_1", "chair_2", "chair_3"]
    max_distance = 0
    farthest_chair = chairs[0]
    for chair in chairs:
        distance = get_distance_between(desk, chair)
        if distance > max_distance:
            farthest_chair = chair
    target_area = get_area_to_sit_on(farthest_chair)

    sara = get_character("Sara")
    sara.set_position(target_area)
    sara.set_target_action("drink")
    return [sara]
    \end{lstlisting}
    \vspace{-1em}
    \end{minipage}
    \caption{An example of a programming-structured prompt used in our event parser. The event parser leverages provided spatial reasoning tools, such as the \texttt{get\_distance\_between()} function in this example, to procedurally deduce solutions for given spatial reasoning tasks.}
    \label{fig:pythonic_prompt}
    \vspace{-1em}
\end{figure}

\begin{figure*}[ht!]
\centering
\includegraphics[width=0.9\linewidth]{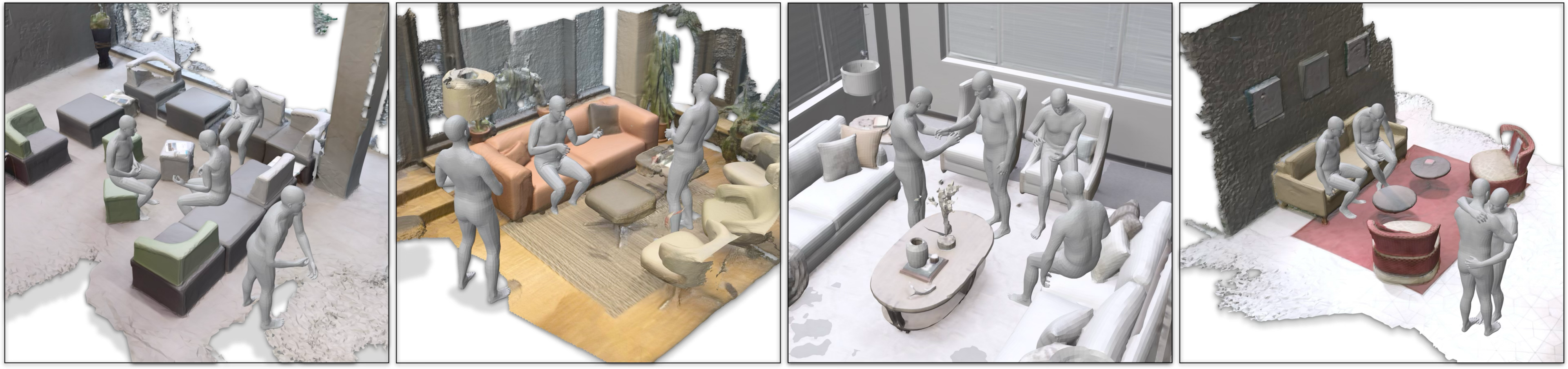}
\caption{Qualitative results of our framework across diverse 3D scenes.}
\label{fig:qualitative_results}
\vspace{-1em}
\end{figure*}

\subsubsection{Event Parser}
The event parser takes the event $e$ generated by the narrator as input and concretizes it into low-level details that the motion synthesis module can digest.
The output event of the narrator is designed to be a short description of the region and the character for the motion, leaving ambiguities on the exact location within the scene. 
The event parser observes the current state of the scene and refines the event description into detailed 3D grounding suitable for motion synthesis.
Specifically, it converts the event description $e$ into $e=(\mathcal{C}_e, \{p_i\}, \{d_i\}, \{a_i\})$, where $\mathcal{C}_e=\{c_i\}$ is the set of involved characters, $p_i\in\mathbb{R}^2$ is the target position, $d_i\in\mathbb{R}^1$ is the target orientation, and $a_i$ is the target action label for each character $c_i$.

Inspired by~\cite{liang2023code, yang2024llm, singh2023progprompt}, we adopt a Python programming structure in the event parser's prompt and utilize spatial reasoning tools that are provided as a form of Python functions.
The programming framework can infer solutions in a well-defined sequence of inquiries, leading to more powerful spatial reasoning capabilities.
Figure~\ref{fig:pythonic_prompt} shows a specific example.
The event parser uses the \texttt{\small get\_distance\_between()} function to deduce spatial details missing from the scene description, such as identifying the farthest seat from a given reference point.
The event parser compiles the output format by calling the \texttt{\small get\_character()} function, and for each character $c_i$, it specifies $p_i$, $d_i$, and $a_i$ using the \texttt{\small set\_position()}, \texttt{\small set\_orientation()}, and \texttt{\small set\_target\_action()} functions, respectively.
If a specific target orientation is not necessary, it can be left unspecified.
When retrieving the positions, the event parser first outputs semantic descriptions using scene objects, and precise coordinates are found afterward to compensate for possible errors of LLMs in coordinate-level reasoning~\cite{davoodi2024llms, mirzadeh2024gsm}.
Specifically, we define a position by its spatial relationship with an anchor object and sample the coordinate representation $p_i$ within the corresponding area. 
Then, if a target orientation is specified, we calculate $d_i$ as the object or character the subject faces after sampling $p_i$.
Otherwise, $d_i$ is set to the character's expected forward direction upon reaching $p_i$.

As with the narrator, the event parser incorporates in-context learning, chain-of-thought reasoning, and self-feedback techniques in its prompt.
Further details about sampling and the full prompts are provided in the supplementary material.

\subsection{Low-level Motion Synthesis Module}
\label{sec:motion_synthesis_module}

After the high-level action planning module provides the narrowed location and character action to generate, the subsequent motion synthesis module concentrates solely on converting the set of events into character motions in a 3D scene in an online manner, detached from the holistic flow of the scene.
Given the positions and actions of individual characters, the motion synthesis may employ any existing framework to reach the location and perform the designated action.
An event is completed and removed from the scene when all characters in $\{c_i\}$ complete their own target actions $a_i$.
We briefly describe our motion synthesis module below, and more details are available in the supplementary material.

We first plan a collision-free trajectory and generate locomotion along it such that the character $c_i$ moves to the target position $p_i$, aligning its facing direction with the target orientation $d_i$.
While reaching the target position $p_i$, the character should avoid collisions with other characters and the surrounding environment.
We first create a 2D grid map that represents navigable areas in the scene, following the approach used in~\cite{wang2022towards}.
On this grid map, our motion synthesis module finds collision-free multi-agent paths using the windowed cooperative $A^*$ algorithm proposed in~\cite{silver2005cooperative}.

After reaching the target position, the motion synthesis module creates motions assigned with the action label $a_i$.
We leverage the motion matching~\cite{clavet2016motion} to quickly generate plausible motions.
If $p_i$ is deliberately set to overlap with an object (e.g., on a chair), the motion synthesis module first makes a transition to a desired posture (e.g., sitting) before creating motions for $a_i$.
There are also group events, where multiple characters are involved, such as chatting or handshake.
Because temporal synchronization is critical, the characters who reach $p_i$ before others maintain an idle motion and wait for the remaining characters. 
Motion synthesis for $a_i$ occurs only after all characters have reached their respective target position $p_i$.

%% file: sec/4_experiment.tex
\section{Experiments}
\label{sec:experiments}
The key functionality of our framework is to perform LLM-based scene-aware multi-agent planning to enable natural interactions and activities of characters within a given 3D scene.
We can successfully populate natural motions of multiple characters within diverse scenes ranging from virtual scenes composed of CAD models (Figure \ref{fig:teaser}) to 3D scans (Figure \ref{fig:qualitative_results}).
More qualitative results are available in the supplementary materials.

To the best of our knowledge, our framework is the first to address this task at this scale; there is no existing baseline for direct comparison with our framework.
Therefore, we propose evaluation criteria to comprehensively assess our framework (Section~\ref{sec:evaluation_criteria}) and create a benchmark designed for this assessment (Section~\ref{sec:benchmark_creation}).
Using the developed benchmark, we evaluate the effectiveness of our proposed key methodologies through an ablation study (Section~\ref{sec:benchmark_results}).
Furthermore, we conduct a user study to visually review and compare the generated results across the different ablation settings (Section~\ref{sec:user_study_results}).

\subsection{Evaluation Criteria}
\label{sec:evaluation_criteria}
We propose a set of criteria to assess several capabilities required for the task as shown below.
%We propose a set of criteria to assess several capabilities required for our task: LLM-based scene-aware multi-agent behavior planning.
% We define each evaluation criterion as following and categorize test cases in the benchmark accordingly.
We define each evaluation criterion as a tag and use these tags to categorize test cases in the benchmark.
% From this point, we define each evaluation criterion as a tag and use these tags to categorize test cases in the benchmark.

\begin{itemize}
  % \item \textbf{Position inference (\textit{PI})}: We evaluate whether the system determines the precise positions of characters at the coordinate level.
  %We evaluate whether the system ensures precise localization when determining character positions at the coordinate level.
  \item \textbf{Object arrangement reasoning (\textit{OA})}:
  Object arrangement reasoning is essential for identifying objects based on spatial relationships (e.g., the chair farthest from the window) or ensuring that characters interact correctly with their surroundings (e.g., to study at a desk, a character needs to sit on the associated chair).
  % We evaluate whether the system accurately understands the spatial relationships between objects to generate logical and physically feasible behaviors.
  \item \textbf{Regional context reasoning (\textit{RC})}:
  Regional context reasoning is required for generating character behaviors that are well-aligned with the contextual meaning of a space (e.g., cooking in the kitchen rather than in the living room).
  % We evaluate whether the system properly infers the contextual meaning of key areas within the scene to generate contextually plausible behaviors.
  \item \textbf{Scene state reasoning (\textit{SS})}:
  Scene state reasoning requires considering the overall scene flow and the current state of other characters (e.g., if a character is using the coffee machine, another character should wait to use it).
  % We evaluate whether the system can generate a proper subsequence plan based on the planning history and the states of other characters. %produce appropriate behaviors that correctly consider the current states of other characters.
\end{itemize}
In addition to the holistic spatial context defined above, we separately evaluate the low-level performance on \textbf{Position inference (\textit{PI})}. The cases with the tag analyze if the system can find the precise positions of characters at the coordinate level.

% \begin{figure}[t]
%   \centering
%   % \fbox{\rule{0pt}{1.5in} \rule{0.9\linewidth}{0pt}}
%   \includegraphics[width=1.1\linewidth]{fig/test_scenes.pdf}
%   \caption{Test scenes.}
%   \label{fig:test_scenes}
% % \vspace{-1em}
% \end{figure}

\begin{figure}[t]
  \centering
  % \fbox{\rule{0pt}{1.5in} \rule{0.9\linewidth}{0pt}}
  \includegraphics[width=0.95\linewidth]{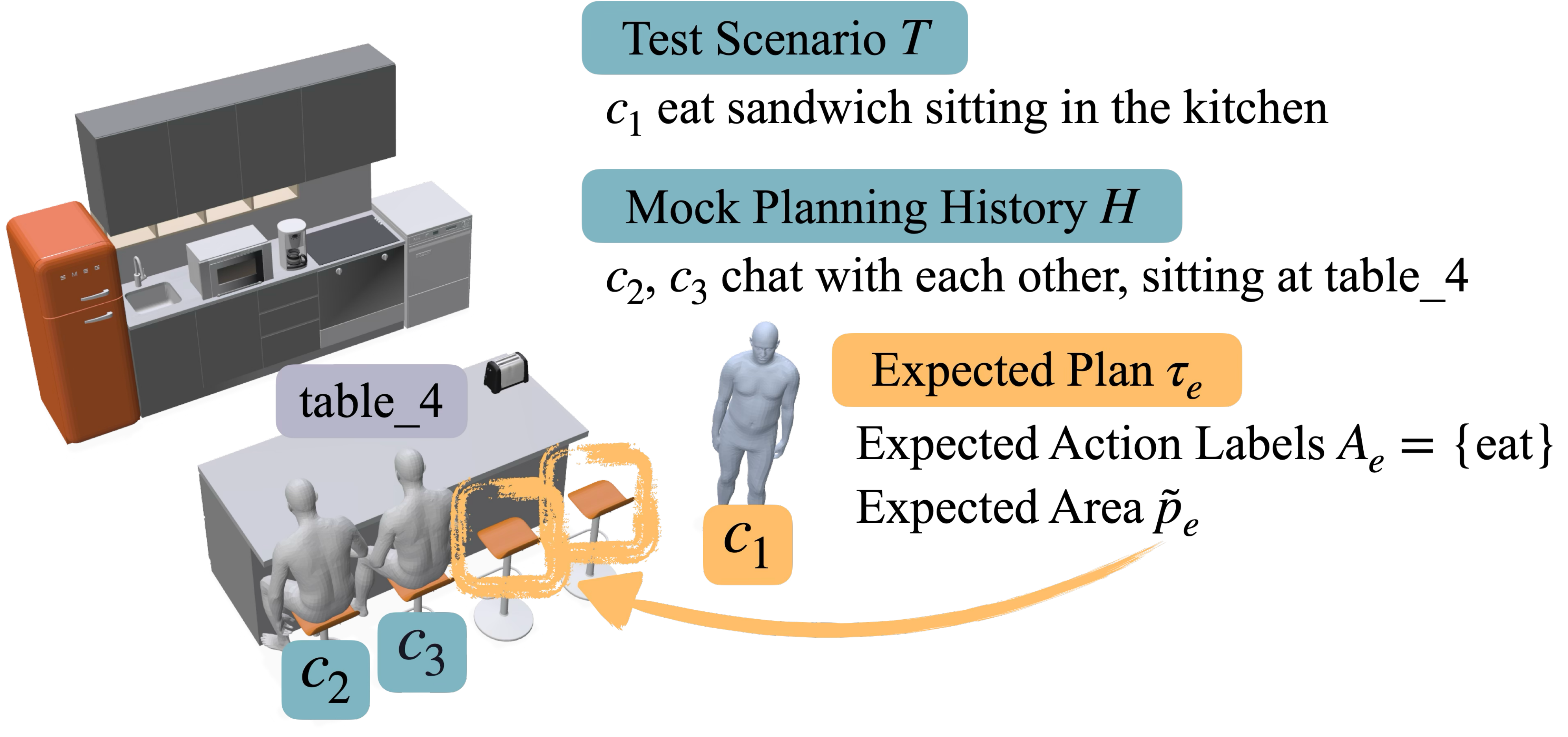}
  \caption{An example of a test case. In this test case, a system is required to identify the chairs in the kitchen and select one that is not currently occupied.}
  \label{fig:test_case}
\vspace{-1em}
\end{figure}

\subsection{Benchmark Creation}
\label{sec:benchmark_creation}
\paragraph{Test Scenes}
% Before the test scenario creation, we first construct test scenes.
% , as illustrated in Figure~\ref{fig:test_scenes}.
We create three scenes---House, Office, and Restaurant---by placing the objects from the HSSD dataset~\cite{khanna2024habitat}.
Each scene is designed to include two to three separate areas with distinct contextual meaning.
To ensure the LLM infers contextual meaning of areas without relying on object labels, we replace several object labels with general ones (e.g., \texttt{\small kitchen\_cabinet} $\rightarrow$ \texttt{\small cabinet}).
We also randomize indices for identically labeled objects to eliminate index-based arrangement cues.
We define a distinct set of available action labels $\mathcal{A}_\mathcal{S}$ for each scene, where $\mathcal{S}\in\{\text{house, office, restaurant}\}$.
This $\mathcal{A}_\mathcal{S}$ is utilized when determining the pass/fail of test cases that use the scene $\mathcal{S}$.

\paragraph{Test Cases}
Our framework can create natural behaviors for multiple characters without any pre-scripted scenarios.
However, for rigorous evaluation, we introduce specific test scenarios in our benchmark.
Each test case consists of a scene $\mathcal{S}$, characters $\mathcal{C}=\{c_1, c_2, \cdots,c_N\}$, test scenario $\mathcal{T}$, mock planning history $\mathcal{H}$, and expected plan $\tau_e=(\mathcal{A}_e, \tilde{p}_e)$.
Here, $N$ is the number of characters, $\mathcal{A}_e$ is the set of expected action labels, and $\tilde{p}_e$ is the expected area where the target character's position is expected to be included.
$\tilde{p}_e$ is expressed through the union or intersection of multiple areas, each defined as a semantic description based on scene objects.
For simplicity, we ignore the character's orientation in the benchmark, and the target character for the expected plan $\tau_e$ is always fixed to the first character $c_1$ in all test cases.
An example of a test case is depicted in Figure~\ref{fig:test_case}.

When running a test case, we input the test scenario $\mathcal{T}$ and mock planning history $\mathcal{H}$ into the system.
The mock planning history $\mathcal{H}$ simulates the prior planning history of the scene, building the necessary context required for the test case.
Then, the system generates the most appropriate next plan $\tau_g=(a_g, \tilde{p}_g\ \text{or}\ p_g)$ by evaluating the given mock planning history and the test scenario.
Here, $\tau_g$ is the generated next plan for the target character, $a_g$ is the planned action label, and $\tilde{p}_g$ or $p_g$ is the planned position.
If the test case evaluates coordinate-level localization, the system outputs a coordinate-level position $p_g$.
Otherwise, it outputs an area-level position $\tilde{p}_g$.
As the final step, the test case's pass/fail is determined by comparing the expected plan $\tau_e$ and the generated plan $\tau_g$.
For a test case to pass, the following conditions must all be satisfied:
\begin{enumerate}
  \item A plan for the target character is generated $(\tau_g\ne\emptyset)$.
  \item The action label is included in both the available action labels and the expected action labels $(a_g\in \mathcal{A_S} \cap \mathcal{A}_e)$.
  \item The character's position is within the expected area $(\tilde{p}_g \subseteq \tilde{p}_e$ or $p_g \in \tilde{p}_e)$.
\end{enumerate}

\paragraph{Benchmark Summary}
We construct the benchmark based on the previously defined evaluation criteria, namely, tags.
Our benchmark consists of 40 test cases in total.
10 test cases are designed for the \textit{PI} tag, while the remaining 30 test cases involve \textit{OA}, \textit{RC}, and \textit{SS} tags.
The test cases for the \textit{PI} tag deliberately use straightforward scenarios to focus solely on the localization performance.
Among the latter 30 test cases, half incorporate two tags for increased complexity.
Additionally, for the latter 30 test cases, we vary the number of characters in the scene and the size of the mock planning history across three levels within the same scenario.
This expands the number of effective test cases and allows us to evaluate the system's scalability by assessing its robustness against increasing agent counts and planning history complexity.
We design our benchmark to ensure that each tag is evenly represented, preventing bias toward any specific evaluation factor.

\paragraph{Evaluation Metrics}
Our benchmark employs two evaluation metrics: success rate and execution rate.
The success rate measures the proportion of passed runs among all test case runs.
On the other hand, the execution rate measures the proportion of runs that are executed without any runtime errors, which are included in the parenthesis next to the success rate in tables.
Runtime errors typically occur when the LLM module references a nonexistent object or generates a response with incorrect syntax.
Any test case run with a runtime error is automatically marked as failed.

\input{tab/benchmark}

\subsection{Benchmark Results}
\label{sec:benchmark_results}
We use GPT-4o~\cite{hurst2024gpt}, Llama-3.1~\cite{dubey2024llama}, and Qwen2.5~\cite{yang2024qwen2} in various sizes as LLM backbones to obtain benchmark results.
The backbones are selected to represent both commercial and open-source modules that are popular and demonstrate competitive performance.
We repeat each test case five times and average the results.

\begin{figure}[t]
  \centering
  \includegraphics[width=0.9\linewidth]{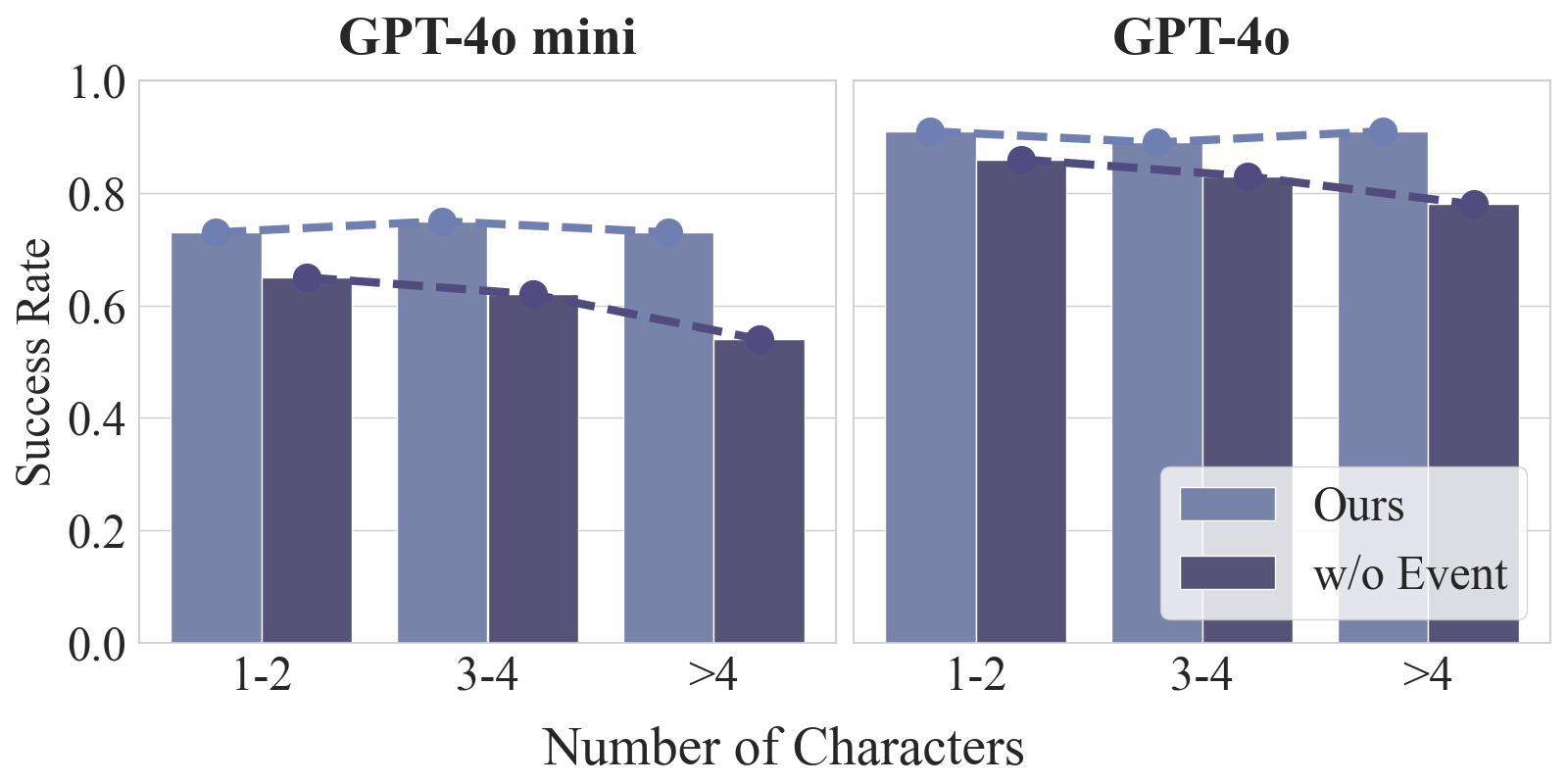}
  \vspace{-1em}
  \caption{Benchmark results by number of characters. Our approach exhibits stable success rate for stories with different number of characters while the performance gap increases as more characters are involved, compared to the version without event representation.}
  \label{fig:scalability}
    \vspace{-1em}
\end{figure}

\begin{figure}[t]
  \centering
  \includegraphics[width=0.9\linewidth]{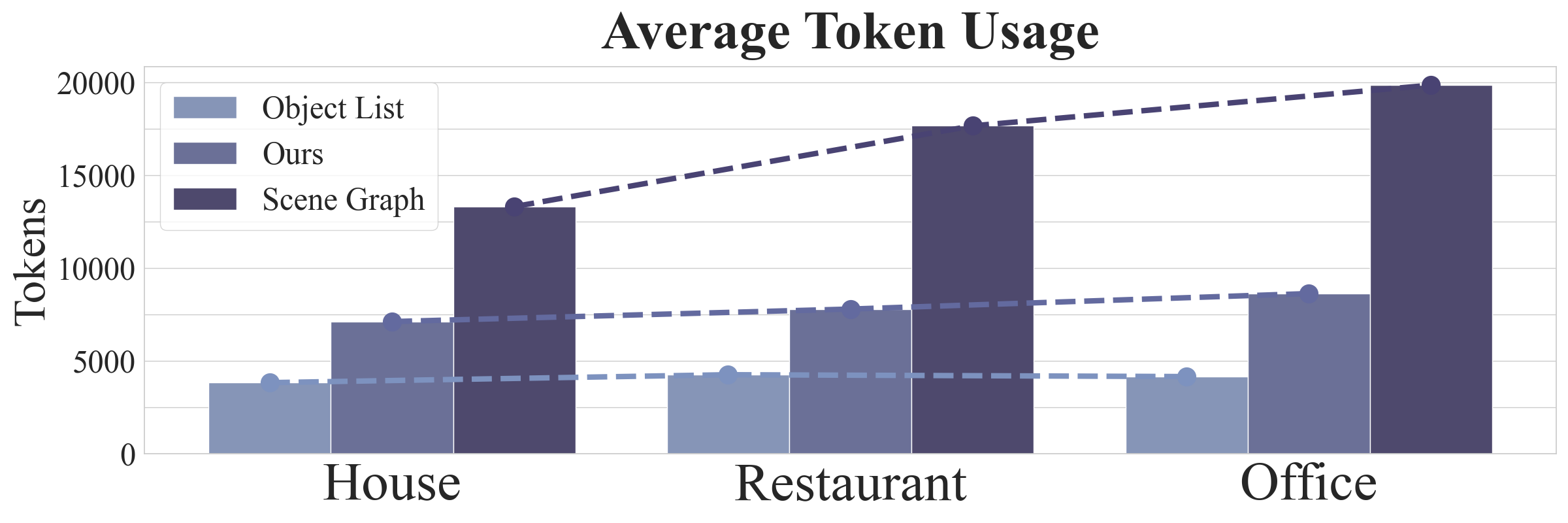}
  \vspace{-1em}
  \caption{Comparison of average token usage. Object List uses the least number of tokens, and Scene Graph uses significantly more number of tokens, especially when the scene complexity increases (Office).}
  \label{fig:token_usage}
  \vspace{-1em}
\end{figure}

\vspace{-1em}
\paragraph{Event-Driven Planning}
Table~\ref{tab:benchmark} presents the benchmark results for test cases with \textit{OA}, \textit{RC}, and \textit{SS} tags.
Each column represents the average score for test cases that include the corresponding tag and each cell displays the success rate, with the execution rate shown in parentheses.
The ablation without event-driven planning (\textit{w/o Event}) handles the entire process with a single unified LLM module without using high-level event representations.
%In Table~\ref{tab:benchmark}, \textit{w/o Event} refers to an ablation setting where our proposed event-based planning pipeline is not used.
%In this setting, the entire planning process is handled by a single unified LLM module without using high-level event representations.
Therefore, all input and output information, except for the scene description and test scenario, is represented in a low-level structured format.
% On the other hand, \textit{Object List} and \textit{Scene Graph} are ablation settings that retain our pipeline but modify the scene description method.
% \textit{Object List} represents the scene as a simple list of objects present in the scene, while \textit{Scene Graph} uses the JSON-formatted scene graph representation as the scene description.
%
%
%As shown in Table~\ref{tab:benchmark}, 
% Our results demonstrate the effectiveness of our proposed methods across various LLM models and sizes.
Compared to \textit{w/o Event}, our method benefits from a cascaded planning pipeline that effectively decomposes the entire planning complexity.
This leads to an overall improvement in planning performance.
In particular, the performance gain in \textit{SS} is relatively larger compared to other tags in GPT-4o models.
The \textit{SS} tag requires a comprehensive understanding of the overall planning state of multiple characters in the scene.
We attribute this to the advantages of our event-based representation, which provides a more efficient and intuitive way to capture the overall scene state.
Additionally, as shown in Figure~\ref{fig:scalability}, our planning pipeline proves to be more effective in terms of scalability.
\textit{w/o Event} struggles with planning as the number of characters in the scene increases, whereas our approach maintains stable performance.

\paragraph{Scene Description}
We also analyze the effect of using scene description generated from an LLM module.
% The ablation studies on scene description methods (\textit{Object List}, \textit{Scene Graph}) demonstrate that our approach contributes to more effective planning compared to other description methods.
We create ablation settings that retain our pipeline but modify the scene description method.
% \textit{Object List} and \textit{Scene Graph} are ablation settings that retain our pipeline but modify the scene description method.
\textit{Object List} represents the scene as a simple list of objects present in the scene, while \textit{Scene Graph} uses the JSON-formatted scene graph representation as the scene description.
Our approach contributes to more effective planning compared to other description methods as shown in Table~\ref{tab:benchmark}.
% In addition to the results in Table~\ref{tab:benchmark}, 
Additionally, we present the average token usage for each description method in Figure~\ref{fig:token_usage} to provide a deeper understanding of their differences.
As presented in Figure~\ref{fig:token_usage}, the object list is the most efficient way to represent scene information and has often been used in previous studies such as ~\cite{singh2023progprompt, kannan2024smart, chen2024sitcom}.
% However, since the object list does not capture any spatial details inherent in the scene, it becomes extremely difficult for the LLM to infer object arrangement or spatial layout from the description.
However, since the object list does not have any spatial information, the planning module with the description is ignorant of object arrangement or spatial layout.
As a result, the object list records the lowest scores in our benchmark, which requires sophisticated scene understanding for planning.

Unlike the object list, the scene graph representation enables necessary spatial reasoning by incorporating detailed attributes of objects and spatial relationships between objects.
% Unlike the object list, the JSON-formatted scene graph representation (abbreviated as scene graph hereafter) enables the necessary spatial reasoning by incorporating detailed attributes of objects and spatial relationships between objects in its representation.
However, the representation contains exhaustive detail, including even minor and insignificant ones, without any abstraction or compression.
% However, it captures every object attribute and spatial relationship in exhaustive detail, even including minor and insignificant ones, without any abstraction or compression.
This inefficiency leads to excessive token consumption (Figure~\ref{fig:token_usage}) and can even interfere with the LLM’s reasoning due to unnecessary details (Table~\ref{tab:benchmark}).
The token inefficiency becomes more pronounced as the scene size and complexity increase.
As shown in Figure~\ref{fig:token_usage}, when comparing the relatively simple `House' scene to the more complex `Office' scene, the scene graph exhibits a significantly larger increase in token consumption compared to alternative approaches.
Compared to the scene graph, our approach allows for a more efficient description of the given scene while preserving key spatial information.

\vspace{-1em}
\paragraph{Positional Inference (PI)}
In Table~\ref{tab:benchmark_pi}, we demonstrate the effectiveness of our area-conditioned position sampling method based on the results for the \textit{PI}-tagged test cases. \textit{Direct Inference} is an ablation setting where the same planning pipeline is used, but the LLM directly outputs coordinates when determining character positions. Although all test cases with the \textit{PI} tag are intentionally designed to be very simple, % \textit{Direct Inference} struggles to accurately determine character positions at the coordinate level across all LLM models.
all LLM models with \textit{Direct Inference} struggle to accurately determine the coordinates of character positions.
In contrast, our system allows the LLM to process information semantically, and obtains precise locations through the proposed area-conditioned position sampling technique.
\input{tab/benchmark_pi}

\subsection{User Study Results}
\label{sec:user_study_results}
We also validate through a user study that our approach represents the given scenario more effectively than other ablation settings.
For the user study, we use a total of four test scenarios, each set in a different scene.
The scenes include House, Office, and Restaurant from the benchmark, as well as MPH11 from the PROX dataset~\cite{hassan2019resolving}.
Compared to the other scenes, MPH11 is relatively simple in both scale and complexity.
Given its limited space, we use a simple scenario containing only two events for the MPH11.
In contrast, for other test scenes, we use more complex scenarios that include 4-5 events.
For each test scenario, users are asked to visually compare results generated from different ablation settings and select the one that best represents the given scenario.
If multiple results are perceived as equally effective, users are allowed to choose more than one.
We generate the results for the user study using GPT-4o mini with a temperature setting of 0.0.

\begin{figure}[t]
    \centering
    \includegraphics[width=0.9\linewidth]{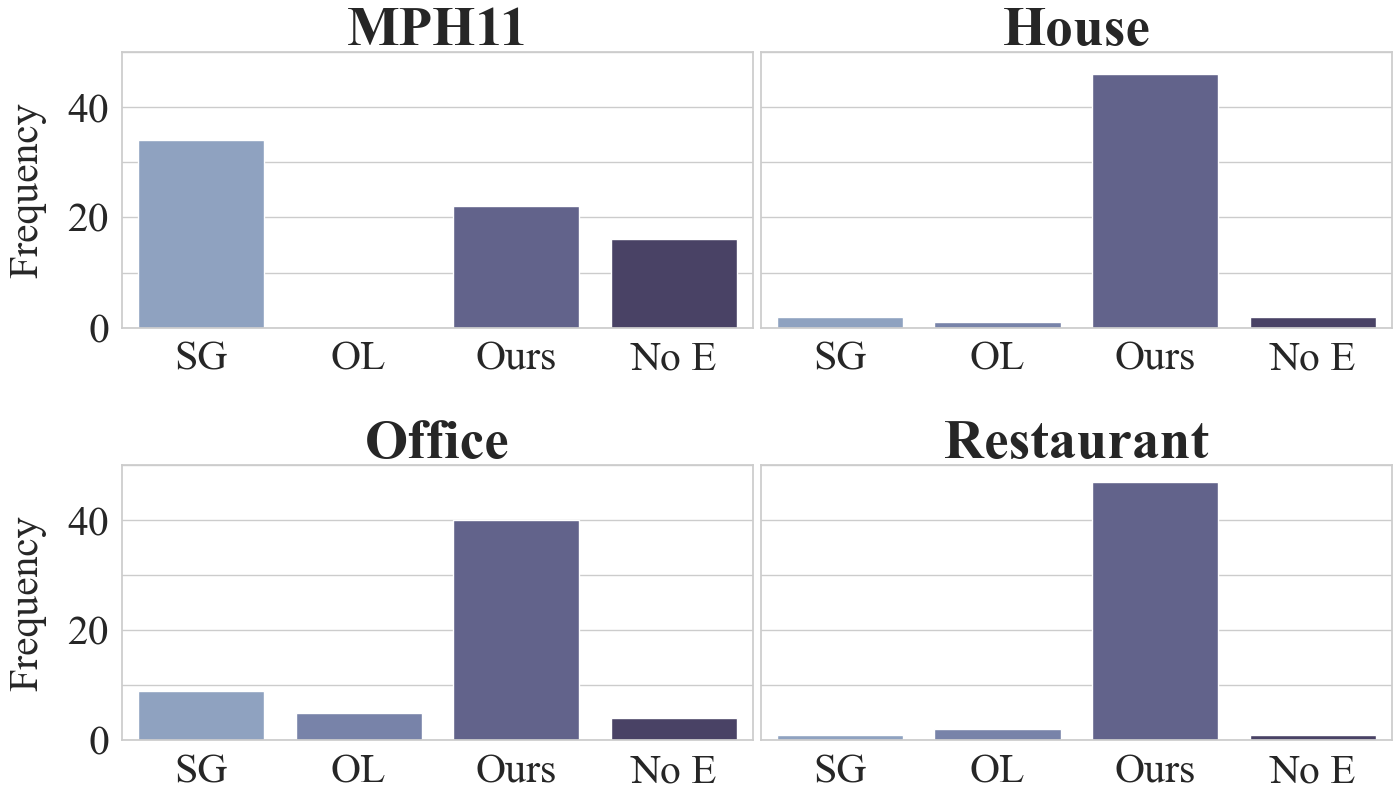}
    \vspace{-1em}
    \caption{User study results. Our system correctly generates the dynamic story line with multiple characters in large scenes, significantly outperforming other baselines, except for simple scenarios with MPH11.}
    \label{fig:user_study_results}
    \vspace{-1em}
\end{figure}

We collected responses from 50 general participants.
The responses to the user study are summarized in Figure~\ref{fig:user_study_results}.
The chart labels SF, OL, and No E correspond to the same ablation settings used in the benchmark: \textit{Scene Graph}, \textit{Object List}, and \textit{w/o Event}, respectively.
In three of our scenarios, our approach received the most selections by a large margin.
Furthermore, the noticeable difference in selection distribution between the MPH11 scene, which involves a relatively smaller-scale problem, and the other scenes suggests that our approach contributes to building a more scalable system.
We provide the test scenarios and visualization of generated results in the supplementary material.

%% file: tab/benchmark.tex
\begin{table*}
    \small
    \centering    
    \resizebox{1.0\textwidth}{!}{
        \begin{tabular}{l|cccc|cccc|cccc}
        \toprule
        \rowcolor[gray]{0.9} 
        Model & \multicolumn{4}{c|}{GPT-4o} & \multicolumn{4}{c|}{GPT-4o mini} & \multicolumn{4}{c}{Llama-3.1-70B} \\
        \toprule
        Metrics & Total & OA & RC & SS & Total & OA & RC & SS & Total & OA & RC & SS \\
        \midrule
        \textit{Ours}
        & \textbf{0.9 (0.98)} & \textbf{0.93 (0.99)} & \textbf{0.9 (0.98)} & \textbf{0.92 (0.98)}
        & \textbf{0.74 (0.96)} & \textbf{0.72 (0.95)} & \textbf{0.72 (0.97)} & \textbf{0.78 (0.93)}        
        & \textbf{0.72 (0.87)} & \textbf{0.78 (0.91)} & \underline{0.76 (0.87)} & \underline{0.61 (0.81)}\\
        
        \textit{w/o Event}
        & \underline{0.82 (0.92)} & \underline{0.88 (0.92)} & \underline{0.86 (0.93)} & 0.77 (0.92)
        & \underline{0.6 (0.95)} & \underline{0.56 (0.99)} & \underline{0.6 (0.92)} & \underline{0.54 (0.92)}        
        & \underline{0.6 (0.81)} & 0.6 (0.78) & 0.62 (0.86) & 0.54 (0.76)\\
        
        \textit{Object List}
        & 0.51 (0.85) & 0.61 (0.78) & 0.28 (0.88) & 0.65 (0.97)
        & 0.34 (0.88) & 0.37 (0.88) & 0.12 (0.86) & 0.51 (0.9)        
        & 0.35 (0.68) & 0.31 (0.7) & 0.29 (0.61) & 0.41 (0.69)\\
        
        \textit{Scene Graph}
        & \underline{0.82 (0.96)} & 0.8 (0.96) & 0.82 (0.94) & \underline{0.87 (0.95)}
        & 0.49 (0.9) & 0.52 (0.93) & 0.32 (0.9) & 0.51 (0.83)        
        & \textbf{0.72 (0.92)} & \underline{0.69 (0.91)} & \textbf{0.77 (0.94)} & \textbf{0.64 (0.85)}\\
        \midrule

        \rowcolor[gray]{0.9} 
        Model & \multicolumn{4}{c|}{Llama-3.1-8B} & \multicolumn{4}{c|}{Qwen2.5-72B} & \multicolumn{4}{c}{Qwen2.5-7B} \\
        \toprule
        Metrics & Total & OA & RC & SS & Total & OA & RC & SS & Total & OA & RC & SS \\
        \midrule
        \textit{Ours}
        & \textbf{0.35 (0.74)} & \textbf{0.37 (0.8)} & \textbf{0.37 (0.77)} & \textbf{0.35 (0.67)}
        & \textbf{0.71 (0.9)} & \underline{0.71 (0.91)} & \underline{0.72 (0.95)} & \textbf{0.65 (0.85)}
        & \textbf{0.39 (0.87)} & \underline{0.38 (0.88)} & \textbf{0.36 (0.87)} & \textbf{0.35 (0.85)}\\
        
        \textit{w/o Event}
        & \underline{0.31 (0.58)} & 0.3 (0.57) & 0.32 (0.6) & \underline{0.28 (0.63)}
        & 0.62 (0.85) & \textbf{0.74 (0.85)} & 0.65 (0.9) & \underline{0.59 (0.85)}
        & \underline{0.35 (0.79)} & \textbf{0.41 (0.86)} & \underline{0.32 (0.79)} & \underline{0.32 (0.74)}\\
        
        \textit{Object List}
        & 0.22 (0.58) & 0.26 (0.65) & 0.13 (0.62) & 0.23 (0.44)
        & 0.44 (0.77) & 0.45 (0.77) & 0.36 (0.83) & 0.52 (0.72)
        & 0.14 (0.77) & 0.09 (0.74) & 0.07 (0.77) & 0.24 (0.67)\\
        
        \textit{Scene Graph}
        & 0.29 (0.75) & \underline{0.31 (0.76)} & \underline{0.35 (0.86)} & 0.16 (0.61)
        & \underline{0.69 (0.84)} & 0.69 (0.81) & \textbf{0.77 (0.93)} & 0.55 (0.72)
        & 0.31 (0.84) & 0.27 (0.88) & 0.23 (0.84) & \textbf{0.35 (0.84)}\\
        \bottomrule
        \end{tabular}
    }    
    \caption{Benchmark result for test cases with \textit{object arrangement reasoning (OA)}, \textit{regional context reasoning (RC)}, and \textit{scene state reasoning (SS)} tags. In total, our method achieves the highest success rate across various LLM models and sizes.}
    \label{tab:benchmark}
    \vspace{-1em}
\end{table*}

%% file: tab/benchmark_pi.tex
\begin{small}
\begin{table}
    \centering    
    \resizebox{0.85\columnwidth}{!}{
        \begin{tabular}{l|ccc}
        \toprule
        \rowcolor[gray]{0.9}
        Model & GPT-4o & GPT-4o mini & Llama-3.1-70B  \\
        \midrule
        \textit{Ours} & \textbf{0.94 (1.0)} & \textbf{0.9 (1.0)} & \textbf{0.94 (1.0)}\\
        \textit{Direct Inference} & 0.46 (1.0) & 0.38 (1.0) & 0.28 (1.0)\\
        \midrule

        \rowcolor[gray]{0.9}
        Model & Llama-3.1-8B & Qwen2.5-72B & Qwen2.5-7B  \\
        \midrule
        \textit{Ours} & \textbf{0.86 (0.88)} & \textbf{1.0 (1.0)} & \textbf{0.94 (1.0)} \\
        \textit{Direct Inference} & 0.14 (0.9) & 0.14 (0.48) & 0.1 (0.8) \\
        \bottomrule
        \end{tabular}
    }    
    \caption{Benchmark result for test cases with \textit{position inference} tag (\textit{PI}). Our proposed method can stably infer accurate positions, whereas \textit{Direct Inference} with LLMs often struggles.}
    \label{tab:benchmark_pi}
    \vspace{-1.5em}
\end{table}
\end{small}

%% file: sec/5_conclusion.tex
\section{Conclusion}
\label{sec:conclusion}
In this study, we propose an LLM-based framework for generating multiple lifelike virtual humans within a given 3D scene, dynamically aligning their behaviors with an emergent storyline.
Characters generated by our framework interact contextually with their surroundings and other individuals, enhancing scene realism.
Our framework performs multi-character behavior planning based on semi-narrative events, allowing the LLM to efficiently capture scene progression and group behaviors. The modular design of our planning pipeline decomposes the overall problem, reducing reasoning complexity for the LLM and enabling more scalable planning.
Additionally, we introduce a context-centric scene description to efficiently convey key spatial information and propose area-conditioned position sampling to mitigate the LLM’s weaknesses in numerical reasoning.
To evaluate our system, we construct a benchmark for a comprehensive assessment of LLM-based scene-aware multi-agent behavior planning.
Through benchmark results and a user study, we demonstrate that our proposed methods contribute to building a scalable and effective system for coordinating multi-agent behaviors, while carefully considering the scene.

\paragraph{Acknowledgments}
This work was supported by the IITP grant [RS-2021-II211341, Artificial Intelligence Graduate School Program (Chung-Ang University)], the NRF grant (No. RS-2023-00208197) funded by the Korea government (MSIT), and Creative-Pioneering Researchers Program through Seoul National University.

%% file: sec/6_supp.tex
% \clearpage
% \setcounter{page}{1}
\setcounter{section}{0}
\setcounter{figure}{9}
\setcounter{table}{2}
\def\thesection{\Alph{section}}
% \maketitlesupplementary

% In this supplementary material, we provide additional details not covered in the main text (Section~\ref{sec:supp:detail}).
% We also present further experimental results (Section~\ref{sec:supp:result}) and visualization of our user study (Section~\ref{sec:supp:user_study}).

\section{Further Details}
\label{sec:supp:detail}

\subsection{Runtime Logic}

Algorithm~\ref{alg:supp:runtime_logic} illustrates the high-level runtime logic of our framework.
In the preprocessing stage (Line 4-6), our framework extracts the scene graph $\mathcal{G}$ from the 3D scene $\mathcal{S}$ and generates a scene description $\mathcal{D}$ using the scene describer.
The generated scene description $\mathcal{D}$ is then used repeatedly by the narrator and event parser during runtime.
In the main runtime loop (Line 7-22), the framework first checks if a new event is required.
A new event is created only when there are characters that are not assigned to any ongoing event.
If a new event is required (Line 9-13), the narrator generates a new event by determining who should be involved among those characters and what activity they should perform.
The event parser then parses the generated event, and our framework assigns the event and its parsed information (target position $p_i$, orientation $d_i$, and action label $a_i$) to the characters involved in the event.
After the behavior planning of the action planning module, the motion synthesis module advances the characters' motions respectively, based on their assigned events (Line 15-21).
If a character is on the move to its target position, the motion synthesis module periodically updates the character's collision-free path to follow using the windowed cooperative $A^*$ algorithm~\cite{silver2005cooperative}.
A character's motion is advanced by synthesizing the next frame of the motion using the motion matching algorithm~\cite{clavet2016motion} based on its current state and assigned action label.
The state of each character is maintained internally to manage the progress of the assigned event and to determine the type of motion to synthesize.
For example, if a character is approaching the target position (\textit{approaching} state), locomotion following the planned path is synthesized.
But if a character is during an interaction after reaching the target position (\textit{interacting} state), a corresponding interaction motion is synthesized according to the assigned action label.
In our framework implementation, we define five states: \textit{idle}, \textit{approaching}, \textit{interacting}, \textit{transition\_in} (standing to sitting), and \textit{transition\_out} (sitting to standing).

\begin{algorithm}[h]
\caption{High-Level Runtime Logic of the Framework}
\label{alg:supp:runtime_logic}
\begin{algorithmic}[1]
\State \textbf{Required:} 3D scene $\mathcal{S}$, characters $\mathcal{C}$
\State \textbf{Optional:} user instructions $\mathcal{T}$

\State
\State Create the 2D grid map of $\mathcal{S}$
\State Extract the scene graph $\mathcal{G}$ from $\mathcal{S}$
\State Generate a scene description $\mathcal{D}$ \Comment{scene describer}
\While{Framework is running}
\State Check if a new event is required
\If{A new event is required}
    \State Generate a new event \Comment{narrator}
    \State Parse the event \Comment{event parser}
    \State Allocate the event to the associated characters
\EndIf

\State
\For{Each character $c_i$ in the scene}
\If{$c_i$ is on the move to the target position}
  \State Update $c_i$'s collision-free path
  \EndIf
\State Advance $c_i$'s motion
\State Update $c_i$'s state
\EndFor
\EndWhile
\end{algorithmic}
\end{algorithm}

\begin{figure}[h]
  \centering
  \includegraphics[width=0.9\linewidth]{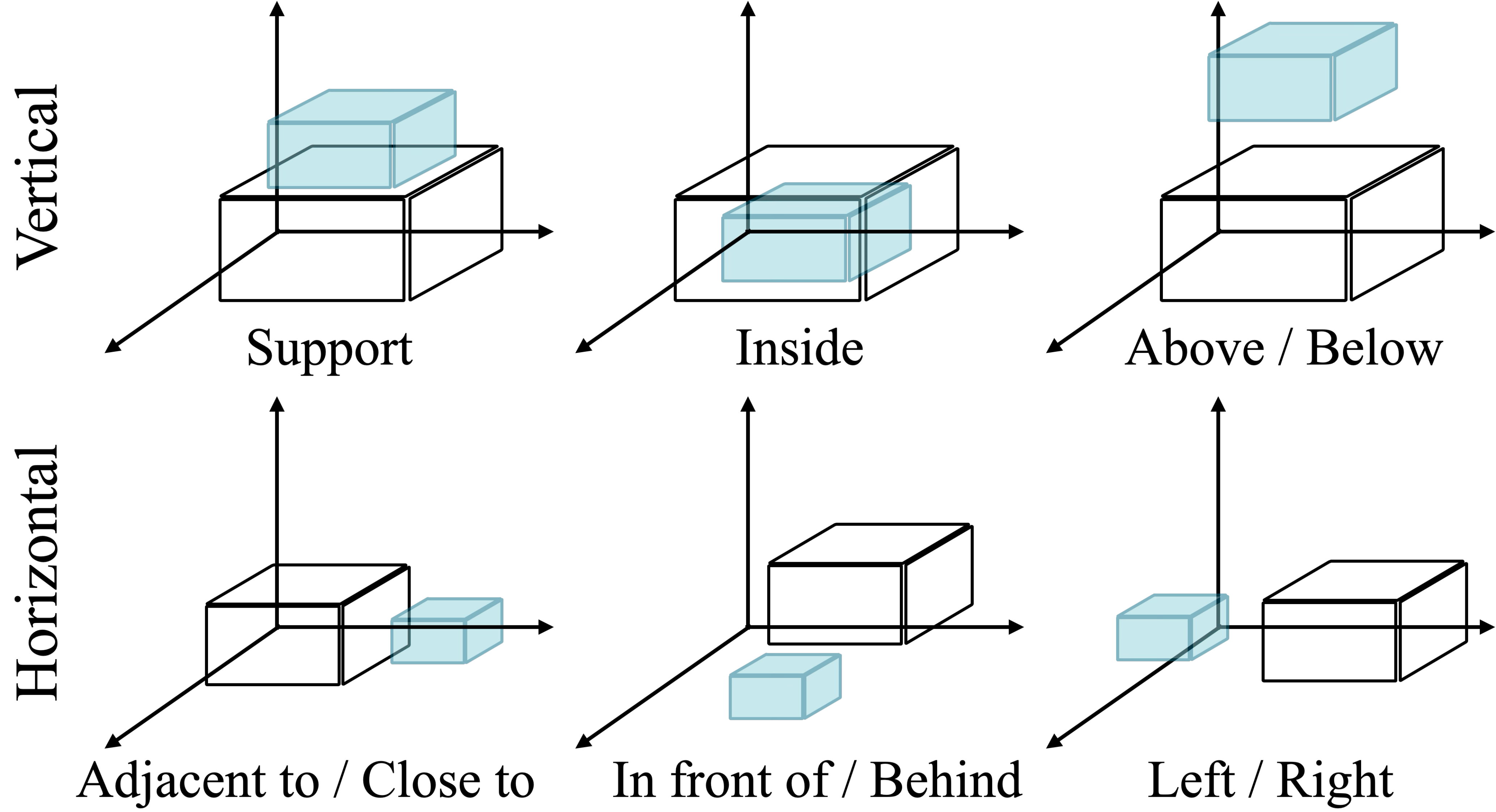}
  \caption{Spatial relationships used in the scene graph construction.}
  \label{fig:supp:spatial_relationships}
\vspace{-1em}
\end{figure}

\subsection{Scene Graph Construction}
In the preprocessing stage, the scene describer (Section~\ref{sec:scene_describer}) generates a textual description $\mathcal{D}$ of the scene $\mathcal{S}$ based on the 3D scene graph.
In this section, we detail the construction of the scene graph below.

To construct a scene graph $\mathcal G = (\mathcal V, \mathcal E)$ from the segmented objects in the 3D scene $\mathcal S$, we follow the automated scene graph construction pipeline proposed in~\cite{jia2024sceneverse}, but with a more simplified list of spatial relationships.
At first, we initialize the nodes $\mathcal V$ with the segmented objects in the 3D scene.
For each object, we compute the z-axis aligned 3D bounding box $b_i=\{p_i^1, p_i^2,...,p_i^8\}\in\mathbb{R}^{8\times3}$ of the object, where the $p_i^j$ $(j\in\{1,...,8\})$ is a vertex composing the $b_i$, and estimate the orientation $d_i\in\mathbb{R}^2$ using the geometric heuristics proposed in~\cite{tahara2020retargetable}.
After the nodes are initialized, we traverse the nodes and compute their spatial relationships to construct the edges $\mathcal E$.
The spatial relationships are categorized into two types: \textit{vertical} and \textit{horizontal} relationships, the full list of which is provided in Figure~\ref{fig:supp:spatial_relationships}.
To avoid the explosion of the number of edges, we first determine the support level of each object based on the \textit{Support} relationships, and limit the spatial relationships based on the support level.
Starting from the support level zero objects, which are directly supported by the floor, if an object $\mathcal{O}_i$ is supported by another object $\mathcal{O}_j$, the support level of $\mathcal{O}_i$ is defined as the support level of $\mathcal{O}_j$ plus one.
Those objects that are not supported by any other objects are defined as the hangable objects.
We allow the horizontal relationships to be computed only between objects with the same support level, and compute \textit{Above/Below} relationships only for the hangable objects.
All the spatial relationships are heuristically computed based on the relative distances and orientations of the 3D bounding boxes of the objects.
For further details of the spatial relationship computation, please refer to our released code.

\subsection{High-level Action Planning Module}

% \begin{figure}[t]
%   \centering
%   % \fbox{\rule{0pt}{2in} \rule{0.9\linewidth}{0pt}}
%   \includegraphics[width=1.0\linewidth]{fig/scene_describer_2.pdf}
%   \caption{Key area extraction on House scene.}
%   \label{fig:supp:scene_describer_example_2}
% \vspace{-1em}
% \end{figure}

\subsubsection{Scene Describer}
\begin{figure}[h]
  \centering
  \includegraphics[width=0.8\linewidth]{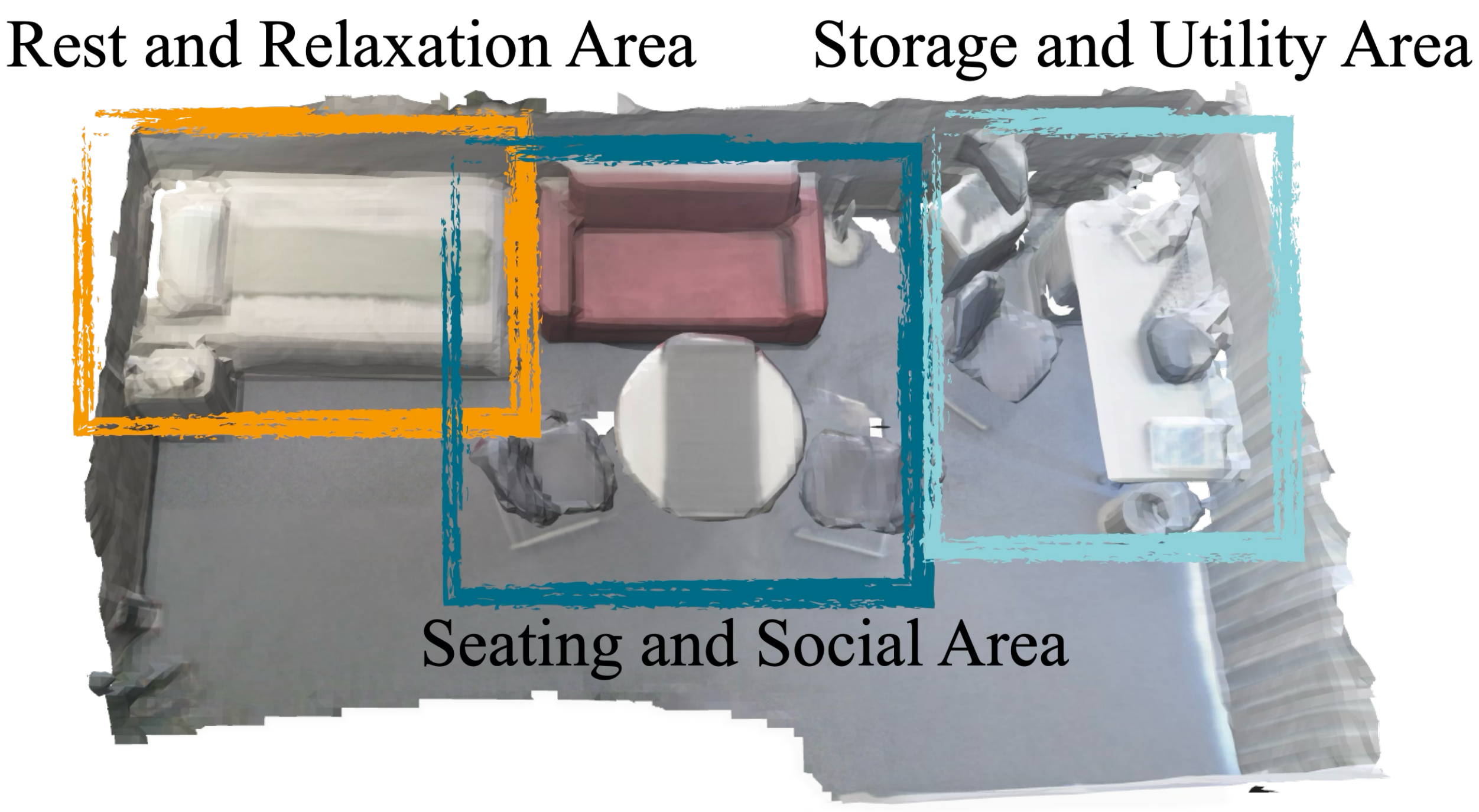}
  \caption{Key area extraction on MPH8 from PROX dataset~\cite{hassan2019resolving}.}
  \label{fig:supp:scene_describer_example_1}
\vspace{-1em}
\end{figure}

In our system, the scene describer takes a scene graph, extracted from the 3D scene and converted into JSON format, as input and transforms it into a context-centric scene description.
We provide object cluster information to help the scene describer better recognize regional context from the given 3D scene.
For object clustering, we apply the DBSCAN algorithm~\cite{ester1996density} to objects present in the scene.
The distance between objects is computed in 3D space as the distance between their bounding boxes.
The key parameters of the DBSCAN algorithm, eps and minimum samples required to form a dense region, are set to 1.0 and 2, respectively.
In Figure~\ref{fig:supp:scene_describer_example_1}, we present examples of how our scene describer extracts key interesting areas from unseen scenes.
% Additionally, we provide the full system prompt used by the scene describer in Figure~\ref{fig:supp:scene_describer}.

\subsubsection{Narrator}
The narrator performs multi-agent behavior planning on a given scene based on semi-narrative events.
The narrator generates new events only for characters not assigned to an `ongoing' event in the current scene.
If the LLM fails to follow this rule correctly, it receives feedback identifying characters that should not be included in the event and regenerates a corrected event based on this feedback.
If all characters are engaged in ongoing events, the narrator does not generate new events and waits until a character completes their event.
% In Figure~\ref{fig:supp:narrator}, we provide the full system prompt used by the narrator.

\begin{figure}[t]
  \centering
  \includegraphics[width=0.8\linewidth]{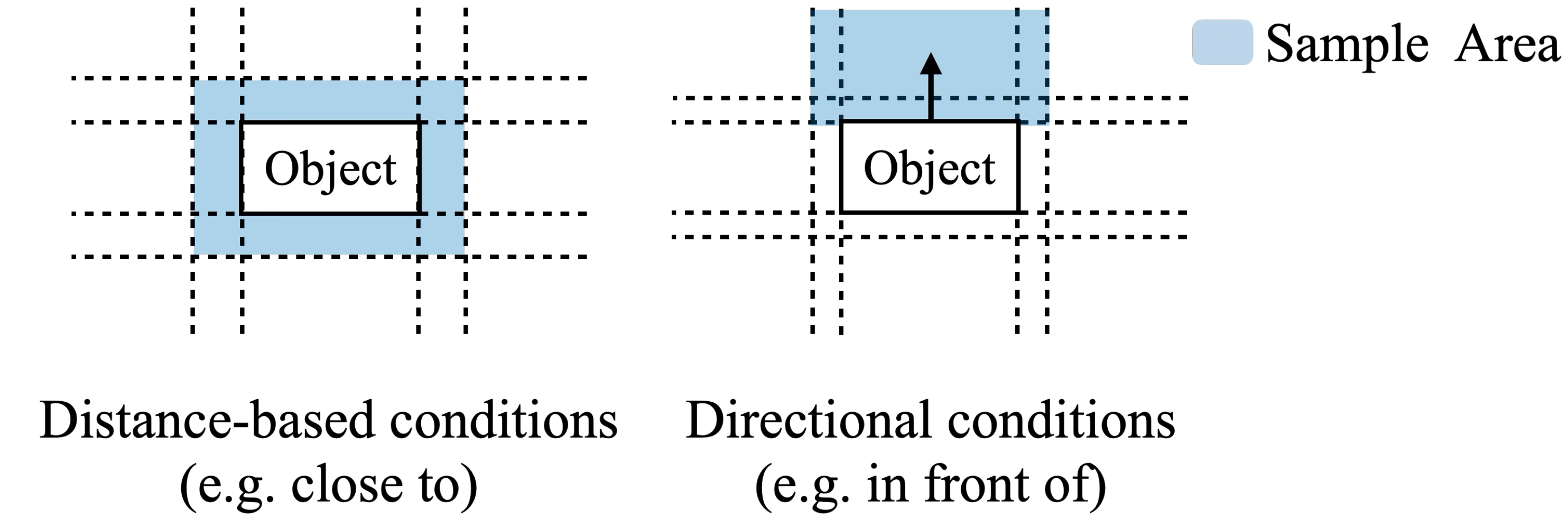}
  \caption{Semantic area representation examples.}
  \label{fig:supp:area}
\vspace{-1em}
\end{figure}

\subsubsection{Event Parser}
The event parser utilizes programming-structured prompts and the area-conditioned position sampling method to parse events into low-level information.
In the programming-structured prompt approach, we enable the event parser to use the following functions as spatial reasoning tools.

\begin{itemize}
    \small
    \item \texttt{get\_object\_supporting(anchor)}
    \item \texttt{get\_objects\_supported\_by(anchor)}
    \item \texttt{get\_objects\_in\_front\_of(anchor)}
    \item \texttt{get\_objects\_behind(anchor)}
    \item \texttt{get\_objects\_left\_of(anchor)}
    \item \texttt{get\_objects\_right\_of(anchor)}
    \item \texttt{get\_objects\_close\_to(anchor)}
    \item \texttt{get\_objects\_associated\_with(anchor)}
    \item \texttt{get\_objects\_between(anchor\_1, anchor\_2)}
    \item \texttt{get\_closest\_object(anchor)}
    \item \texttt{get\_intersected\_area(area\_1, area\_2)}
    \item \texttt{get\_distance\_between(object\_1, object\_2)}
    \item \texttt{is\_object\_occupied(object)}
    \item \texttt{is\_object\_of\_label(object)}
\end{itemize}

Using these functions, the event parser can more easily retrieve objects and determine the appropriate area for character target position sampling based on the retrieved objects.

Our area-conditioned position sampling method enables the LLM to process a character’s target position at a semantic level.
To achieve this, the event parser is provided with the following area retrieval functions.

\begin{itemize}
    \small
    \item \texttt{get\_area\_to\_interact\_with(object)}
    \item \texttt{get\_area\_to\_sit\_on(object)}
    \item \texttt{get\_area\_adjacent\_to(object)}
    \item \texttt{get\_area\_close\_to(object)}
    \item \texttt{get\_area\_in\_front\_of(object)}
    \item \texttt{get\_area\_behind(object)}
    \item \texttt{get\_area\_left\_of(object)}
    \item \texttt{get\_area\_right\_of(object)}
    \item \texttt{get\_area\_between(object\_1, object\_2)}
    \item \texttt{get\_area\_aligned\_with(object\_1, object\_2)}
\end{itemize}

As shown in Figure~\ref{fig:supp:area}, the event parser can meaningfully represent a character’s target position without directly handling coordinate-level representations.
Once an area is specified, the exact coordinates are sampled from within the area.
The specific area size represented by each semantic expression, such as \textit{close to}, is controlled by user hyperparameters.
% In Figure~\ref{fig:supp:event_parser}, we provide the full system prompt used by the event parser.

\subsection{Low-level Motion Synthesis Module}
Our framework requires generating various types of motion to represent characters' daily life activities, including path-following locomotion, human-scene interaction motions, and human-human interaction motions.
To efficiently cover these diverse motion types and generate stable motions in an online manner, we implement the motion synthesis module using the motion matching algorithm~\cite{clavet2016motion}.
Our motion synthesis module utilizes SMPL-X~\cite{pavlakos2019expressive} to represent character bodies and synthesize character animations at a frame rate of 30 fps.

\subsubsection{Motion Database}
Prior to motion synthesis, our framework defines a set of action labels, and we construct separate motion databases corresponding to each action label.
Specifically, motions for daily life activities such as locomotion, drinking, eating, and laptop usage are collected from the AMASS dataset~\cite{mahmood2019amass} and Mixamo~\cite{mixamo}.
Human-scene interaction motions like sitting and lying down are gathered from the SAMP dataset~\cite{hassan2021stochastic}.
Human-human interaction motions, including chatting, hugging, and handshaking, are sourced from the Inter-X dataset~\cite{xu2024inter}.
All motions are downsampled initially to align with the 30 fps.
Each action label has a dedicated motion database, allowing efficient database searching and the use of distinct matching features tailored to the characteristics of each action.

\subsubsection{Motion Matching Details}
\begin{figure}[h]
  \centering
  \includegraphics[width=0.8\linewidth]{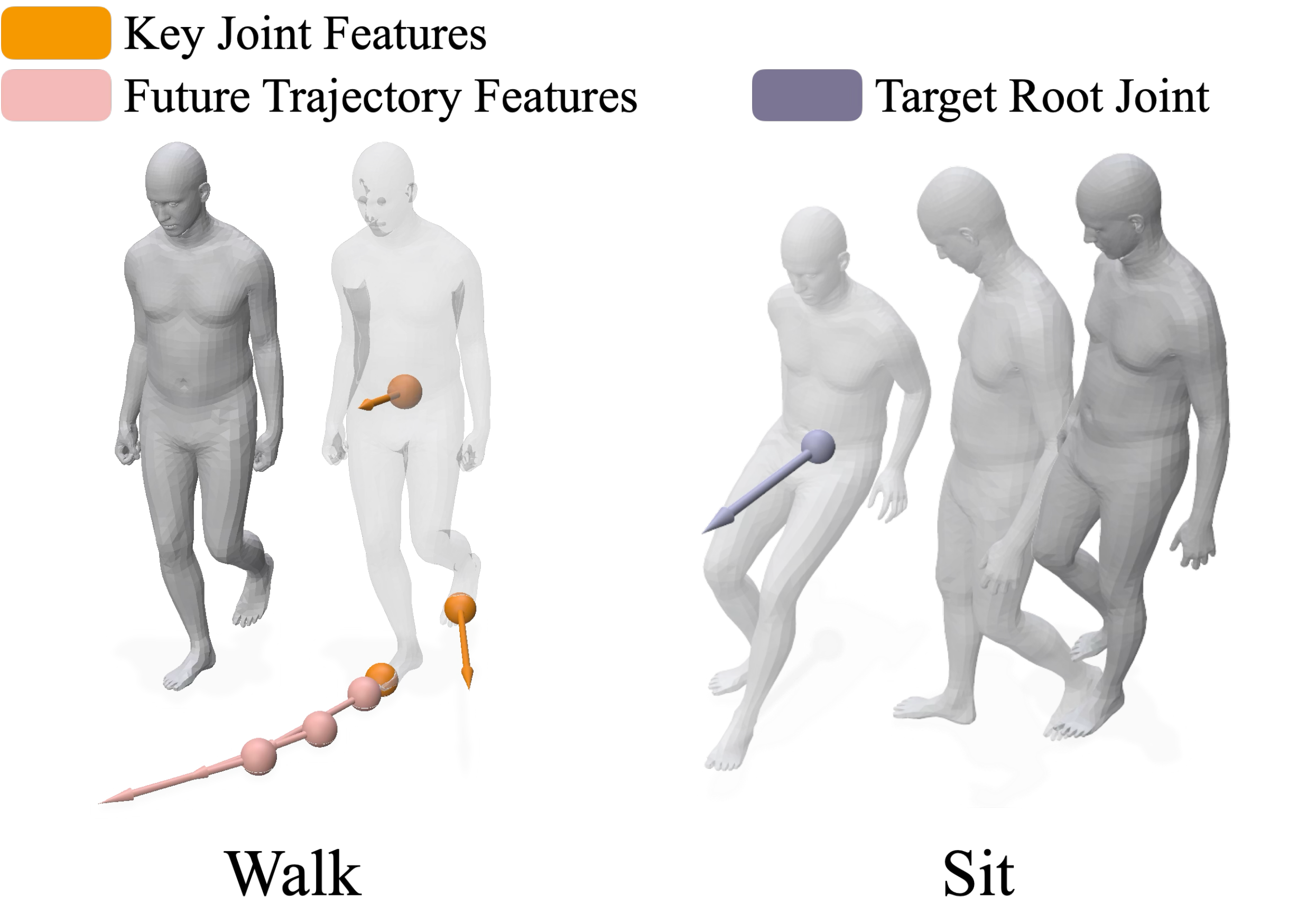}
  \caption{Matching features example.}
  \label{fig:supp:matching_feat}
\vspace{-1em}
\end{figure}

Our motion synthesis module utilizes the following matching features:
\begin{itemize}
\item \textbf{Keyjoint Positions}: Positions of key joints $J$ expressed in the character's local frame ($\mathbb{R}^{3J}$).
\item \textbf{Keyjoint Velocities}: Velocities of key joints $J$ in the local frame ($\mathbb{R}^{3J}$).
\item \textbf{Future Positions}: Ground-projected 2D positions of the future trajectory (at 10, 20, and 30 frames ahead) in the character's local frame ($\mathbb{R}^6$).
\item \textbf{Future Directions}: Ground-projected 2D facing directions of the future trajectory (at 10, 20, and 30 frames ahead) in the character's local frame ($\mathbb{R}^6$).
\item \textbf{Relative Position}: 2D relative position of the character with respect to a specified target position ($\mathbb{R}^2$).
\item \textbf{Relative Velocity}: 2D relative velocity of the character concerning a specified target position ($\mathbb{R}^2$).
\item \textbf{Relative Direction}: 2D relative direction of the character towards a specified target direction ($\mathbb{R}^2$).
\item \textbf{Target Root Height}: Height of the character's target root position ($\mathbb{R}^1$).
\end{itemize}
Keyjoint positions, keyjoint velocities, future positions, and future directions are all represented local to the character's root (pelvis) and facing direction.

As shown in Figure~\ref{fig:supp:matching_feat}, when generating locomotion along a defined path, we employ keyjoint positions, keyjoint velocities, future positions, and future directions as matching features and \texttt{pelvis}, \texttt{spine3}, \texttt{right\_foot} and \texttt{left\_foot} as keyjoints.
For human-scene and human-human interactions, we utilize relative position, velocity, and direction as primary matching features, with an additional target root height feature specifically included for human-scene interactions to ensure accurate sitting positions.
For in-place activities such as eating and drinking, matching relies solely on keyjoint positions and velocities with \texttt{pelvis}, \texttt{spine3}, \texttt{right\_wrist}, \texttt{left\_wrist}, \texttt{right\_foot} and \texttt{left\_foot} as keyjoints.
The pose vector structure and next-frame generation process for actual animation follow the methodologies presented in ~\cite{holden2020learned}.

\subsection{Benchmark}
\subsubsection{Test Scenes}

\begin{figure}[h]
  \centering
  \includegraphics[width=0.8\linewidth]{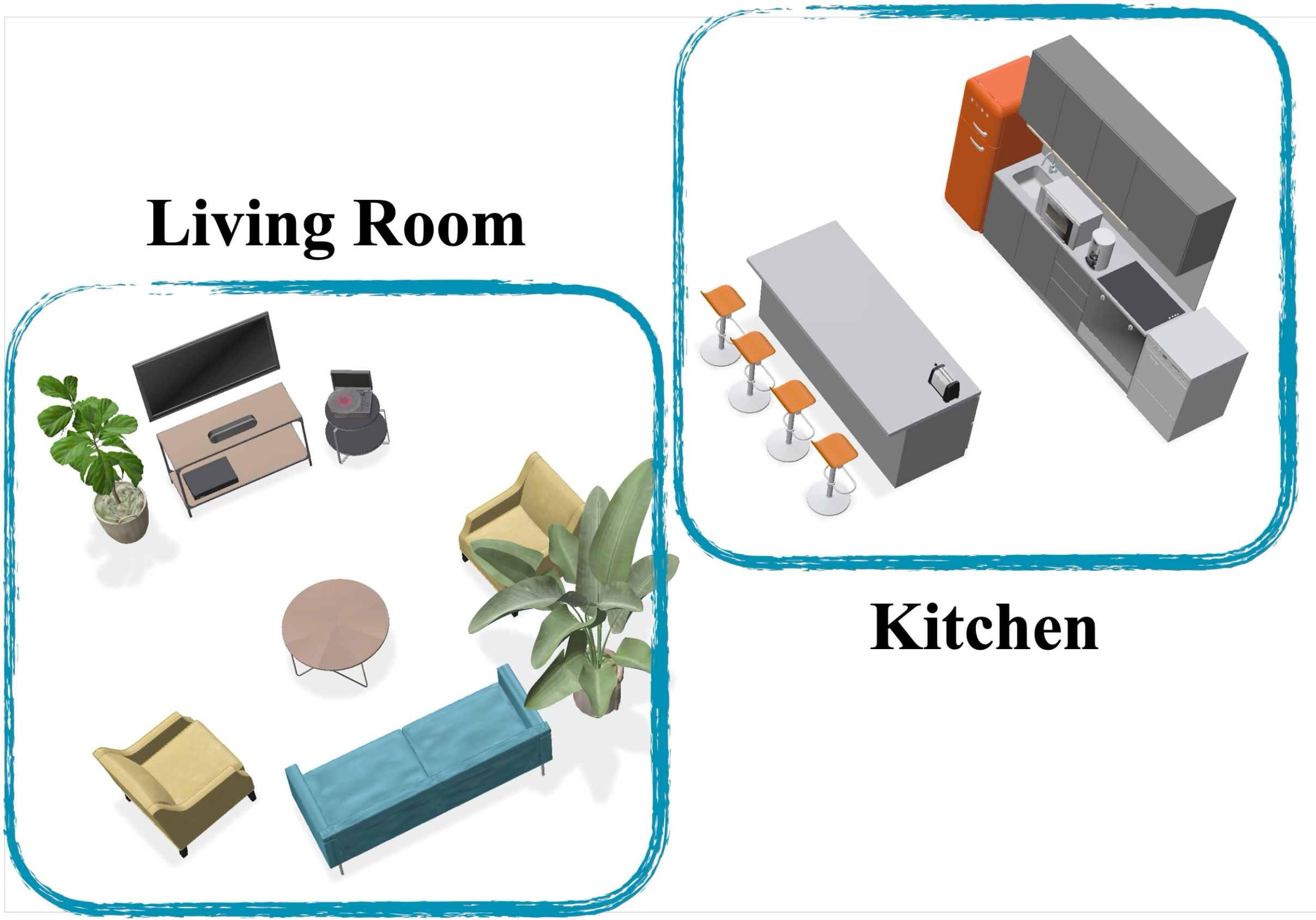}
  \caption{House scene.}
  \label{fig:supp:house}
  \vspace{-1em}
\end{figure}

\begin{figure}[h]
  \centering
  \includegraphics[width=0.9\linewidth]{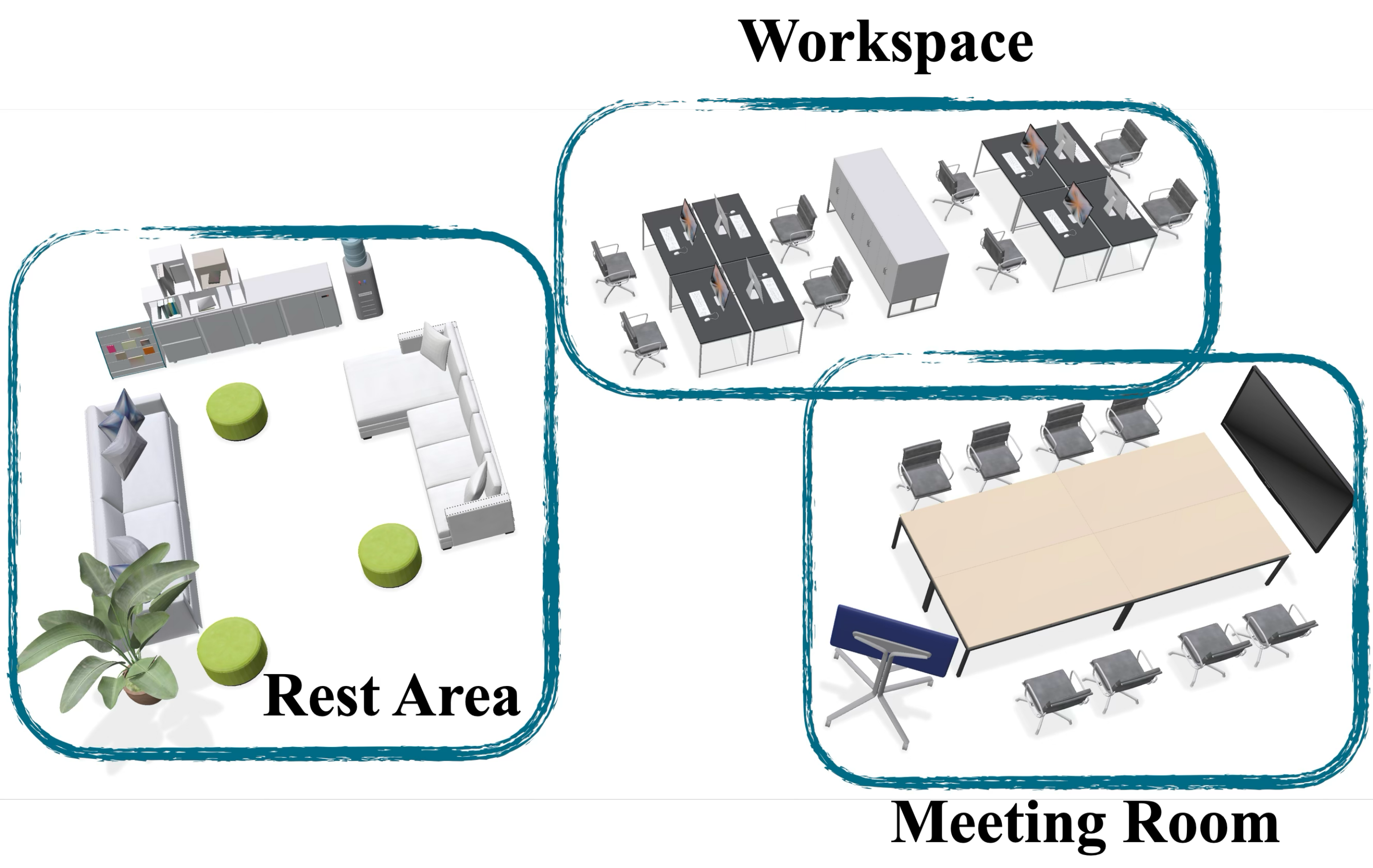}
  \caption{Office scene.}
  \label{fig:supp:office}
  \vspace{-1em}
\end{figure}

\begin{figure}[h]
  \centering
  \includegraphics[width=0.8\linewidth]{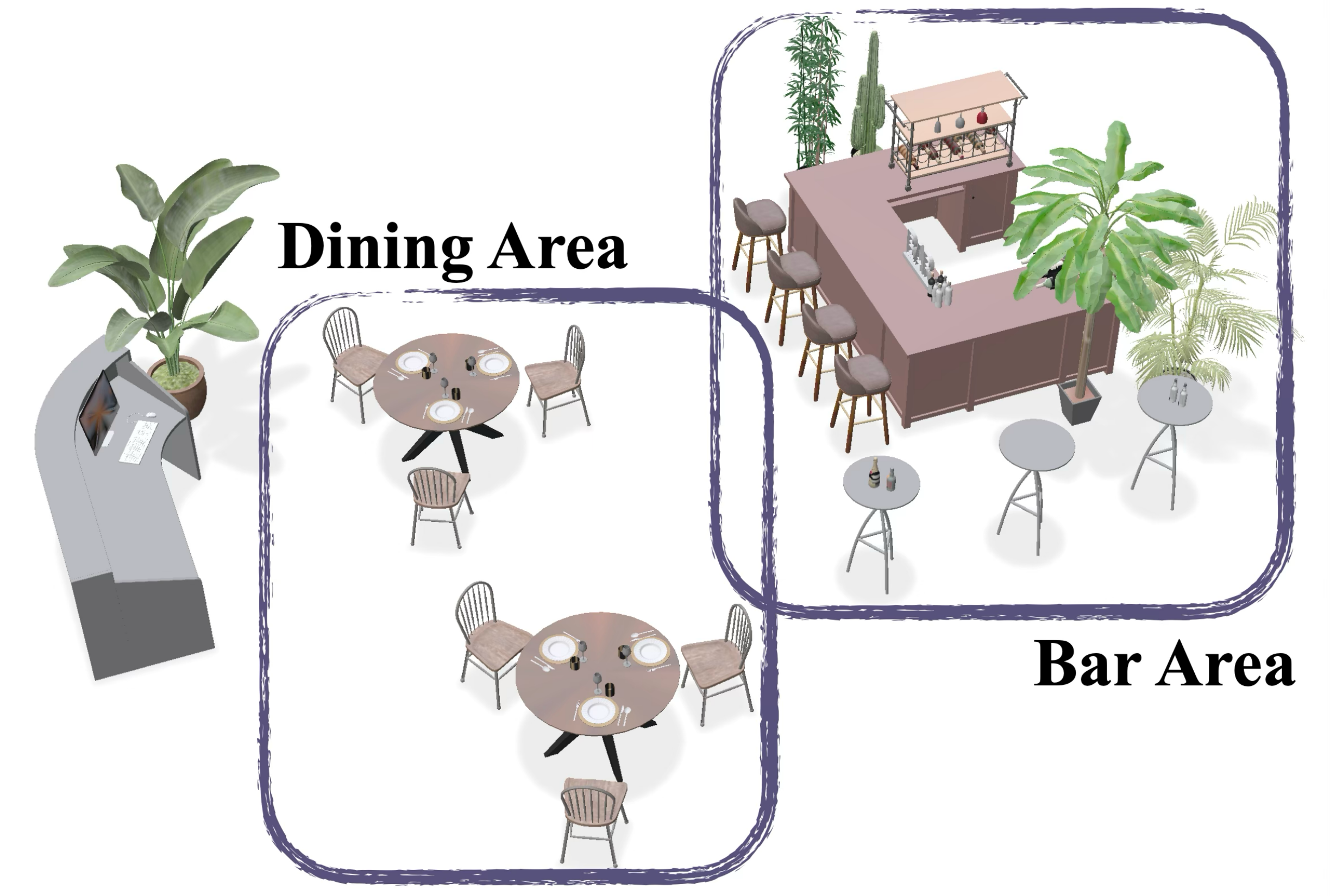}
  \caption{Restaurant scene.}
  \label{fig:supp:restaurant}
  \vspace{-1em}
\end{figure}

In Figure~\ref{fig:supp:house}, \ref{fig:supp:office}, and \ref{fig:supp:restaurant}, we provide visualizations of our test scenes used in our benchmark.
Each scene is designed to include two to three separate areas with distinct contextual meaning.
The House scene is approximately 51.57$m^2$ in size and was created by placing 23 objects from 14 different object categories.
The Office scene is approximately 160.2$m^2$ and includes 51 objects from 11 different object categories.
The Restaurant scene is approximately 72.25$m^2$ and consists of 39 objects from 11 different object categories.

\subsubsection{Test Settings}
Here, we provide additional details of our benchmark test settings.
To address the randomness inherent in LLMs, we repeat each test case five times and average the results.
We also set the temperature parameter, which affects output variability, to 0.1 throughout all experiments.
Such a low temperature value generally makes LLM responses more deterministic.
The scene descriptions are pre-generated in five versions for each test scene using GPT-4o, and the corresponding version with the matching index of trials is fed into the action planning module, such that the $n^\text{th}$ trial observes $n^\text{th}$ scene description.
The LLM planning module also observes input prompts with examples.
We prepare a set of examples for each tag, and the system dynamically selects examples with the tags of the current test case.
None of the examples include test scenes, ensuring that the LLM performs the benchmark in unseen environments.

\section{Additional Results}
\label{sec:supp:result}

\input{tab/benchmark_supp}

\subsection{Additional Benchmark Results}
In Table~\ref{tab:supp:benchmark_supp} we present additional benchmark results that were not included in the main text.
The results from the additional LLM models further confirm that our approach achieves the best overall performance in scene-aware multi-agent planning.

\input{tab/benchmark_vlm}

\subsection{Experiment using Vision-Language Model}

\begin{figure}[h]
  \centering
  \includegraphics[width=0.8\linewidth]{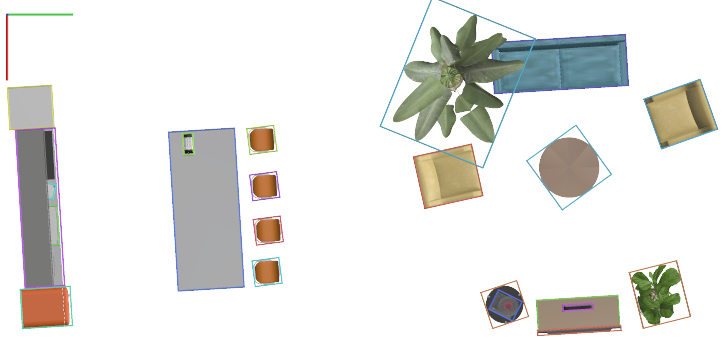}
  \caption{An example of a top-view image used in our VLM-based approaches.}
  \label{fig:supp:house_top_view}
\end{figure}

We additionally conduct experiments on planning methods using vision-language models (VLMs).

\paragraph{\textit{Vision-based Description}}
First, we evaluate how well a VLM utilizes visual information to generate high-quality scene descriptions.
In the \textit{Vision-based Description} approach, we maintain the existing planning pipeline but replace the scene description input for both the narrator and event parser with a vision-generated scene description.
To achieve this, we modify the scene describer, which previously generated descriptions based on scene graphs.
Instead, as shown in Figure~\ref{fig:supp:house_top_view}, the updated scene describer generates detailed descriptions using top-view images along with object labels and position information.

\paragraph{\textit{Vision-based Planning}}
In the \textit{Vision-based Planning} approach, the narrator and event parser perceive scene information through visual inputs rather than textual scene descriptions.
To enable this, we replace the scene description previously provided as input with a top-view image along with object labels and position information, allowing the system to perform planning based on visual data.

\paragraph{Benchmark results for VLM-based approaches}
In Table~\ref{tab:supp:benchmark_vlm}
Table~\ref{tab:supp:benchmark_vlm} presents the benchmark results for the \textit{Vision-based Description} and \textit{Vision-based Planning} approaches described earlier.
For these experiments, we use GPT-4o and GPT-4o Mini as foundation models capable of processing visual information.
The benchmark settings remain the same as in original experiments.

As shown in Table~\ref{tab:supp:benchmark_vlm}, vision-based planning methods perform far worse than our text-based approach.
This suggests that more refined methodologies are needed to achieve effective planning through vision-based scene understanding.
Further exploration in this area could lead to improvements in future work.

\section{User Study}
\label{sec:supp:user_study}
\begin{figure}[h]
  \centering
  \includegraphics[width=0.9\linewidth]{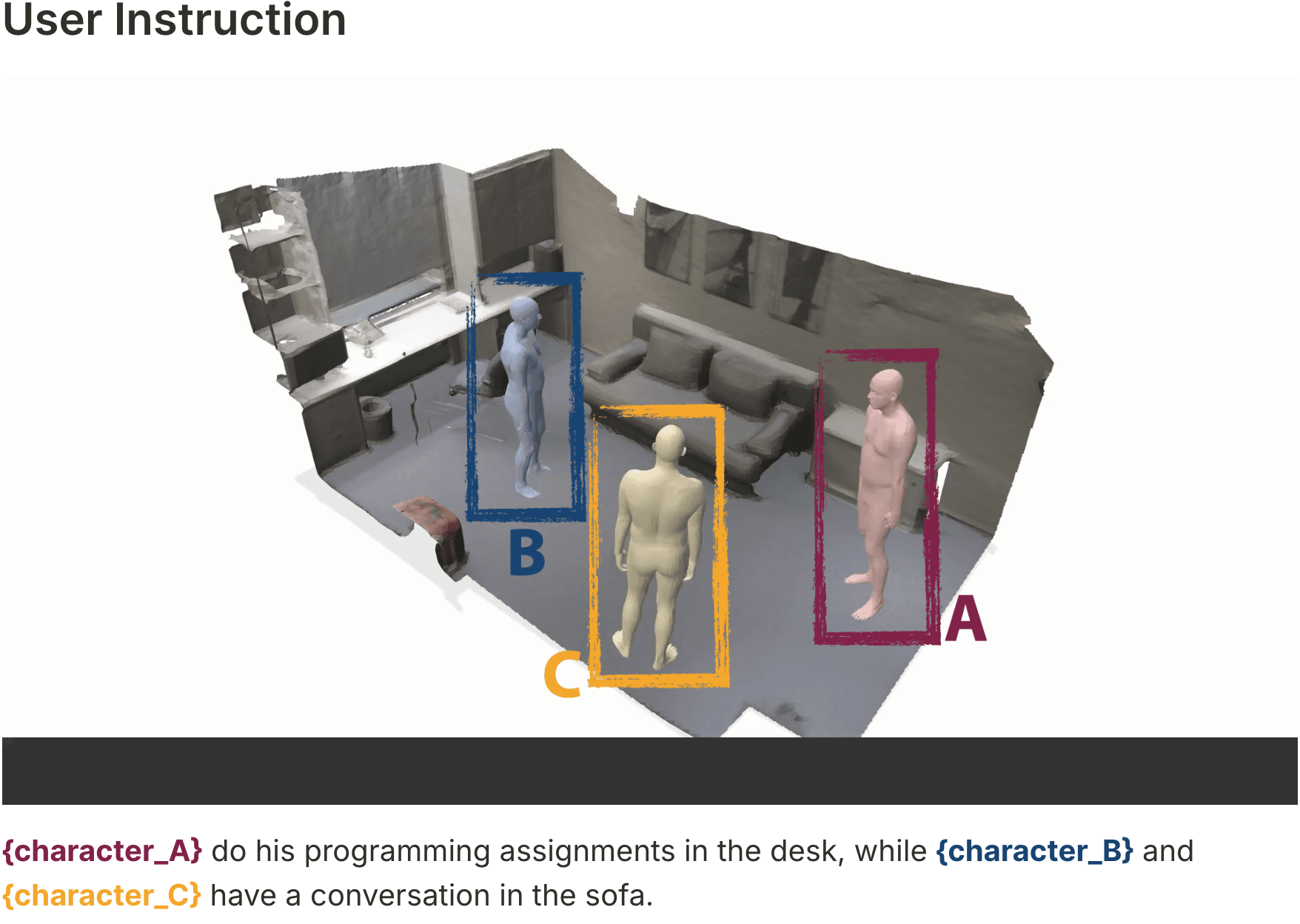}
  \caption{Test scenario employed in the user study for the MPH11 scene.}
  \label{fig:supp:user_study_mpg11}
  \vspace{-1em}
\end{figure}

\begin{figure}[h]
  \centering
  \includegraphics[width=0.9\linewidth]{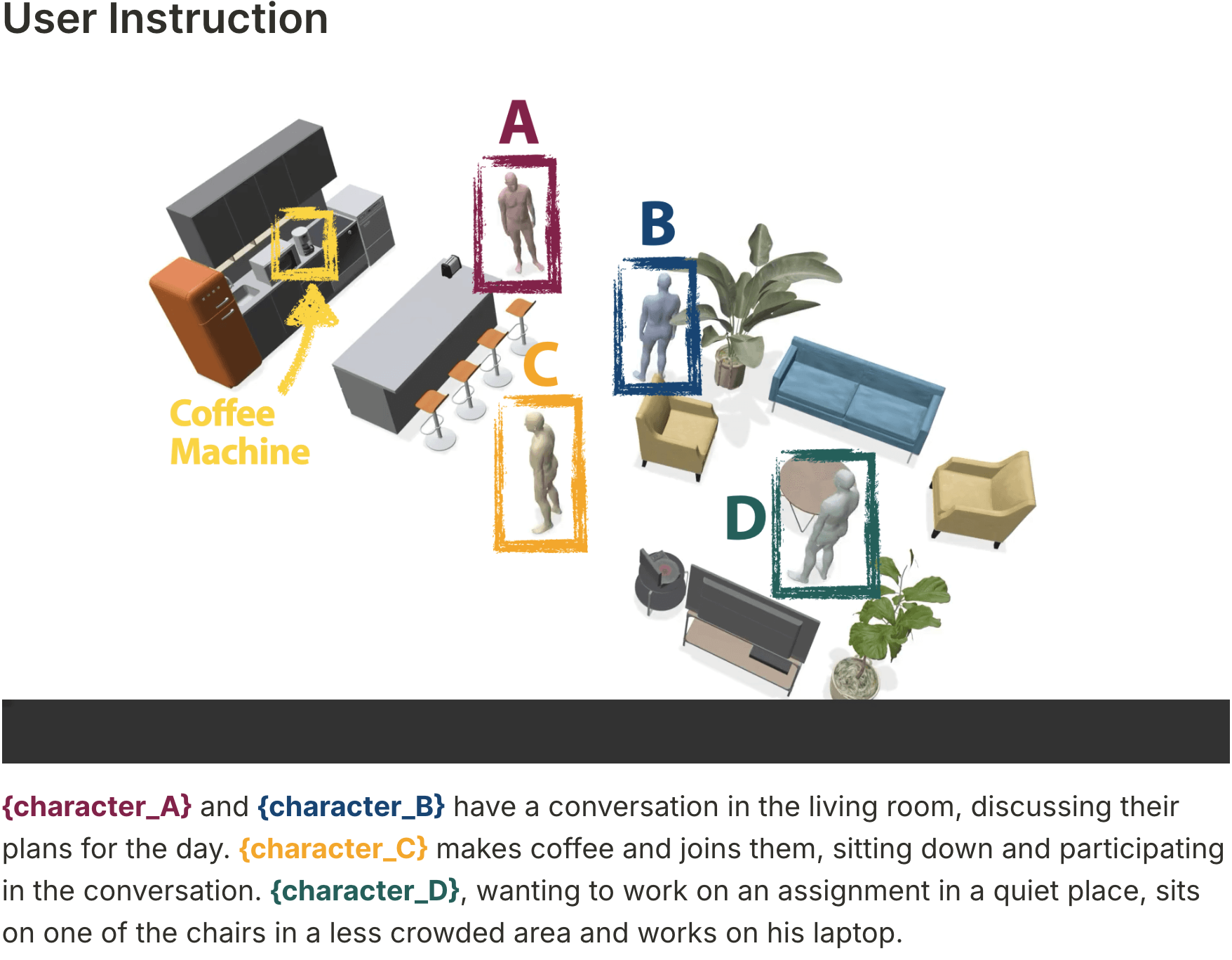}
  \caption{Test scenario employed in the user study for the House scene.}
  \label{fig:supp:user_study_house}
  \vspace{-1em}
\end{figure}

\begin{figure}[h]
  \centering
  \includegraphics[width=0.9\linewidth]{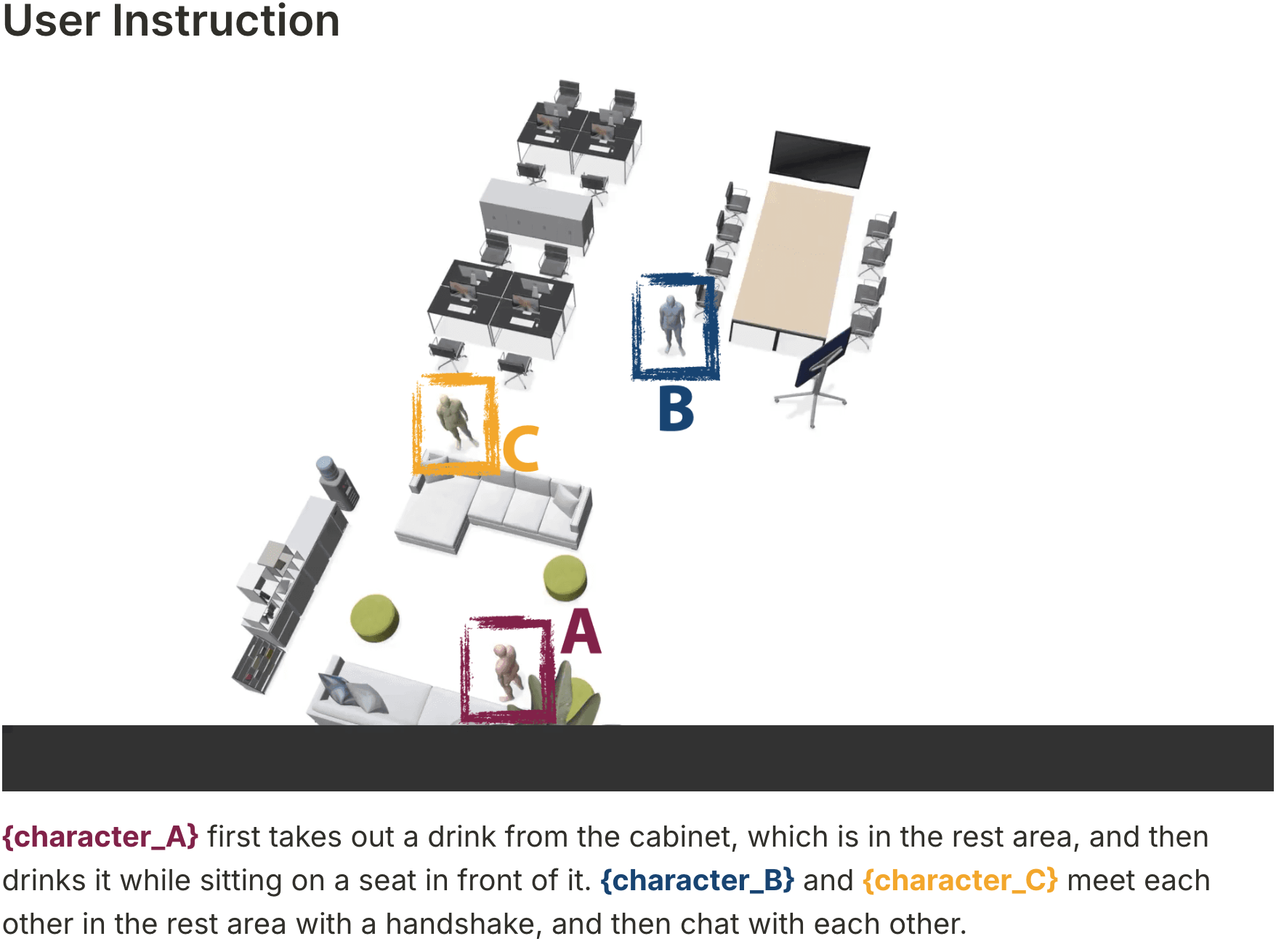}
  \caption{Test scenario employed in the user study for the Office scene.}
  \label{fig:supp:user_study_office}
  \vspace{-1em}
\end{figure}

\begin{figure}[h!]
  \centering
  \includegraphics[width=0.9\linewidth]{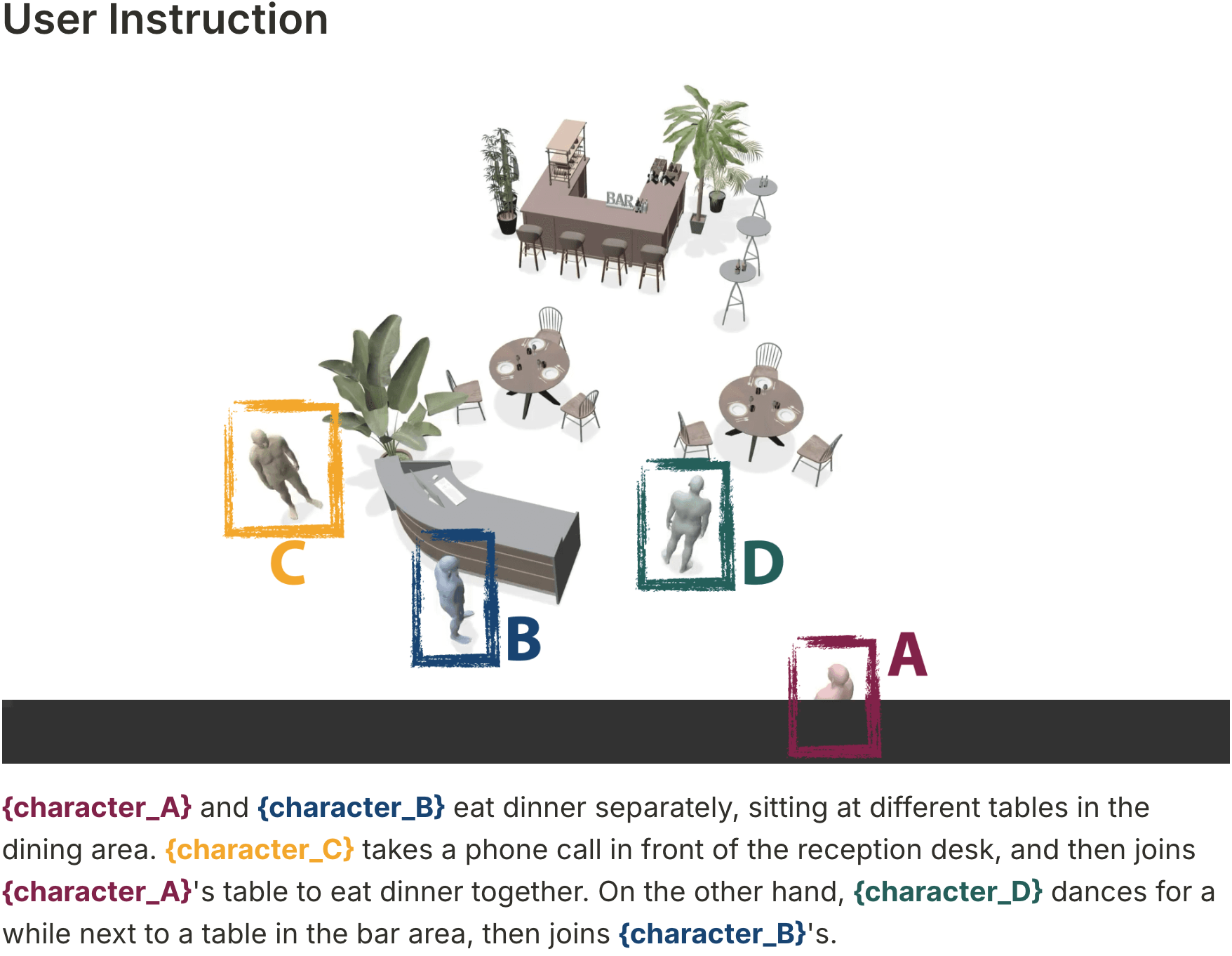}
  \caption{Test scenario employed in the user study for the Restaurant scene.}
  \label{fig:supp:user_study_restaurant}
  \vspace{-1em}
\end{figure}

In this section, we present the test scenarios used in the user study, as shown in Figures~\ref{fig:supp:user_study_mpg11}, \ref{fig:supp:user_study_house}, \ref{fig:supp:user_study_office}, and \ref{fig:supp:user_study_restaurant}.
For each scenario, users are provided with full videos and snapshots of results generated from different ablation settings. They visually examine these results to identify any misrepresented events in the user instruction and ultimately select the outcome they find most accurate.
We provide all full videos used in the user study in the supplementary video.

%% file: tab/benchmark_supp.tex
% Llama-3.1-405B
% 0.66 (0.82) & 0.74 (0.84) & 0.71 (0.88) & 0.52 (0.68) &
% 0.6 (0.71) & 0.46 (0.58) & 0.68 (0.77) & 0.6 (0.72) &
% 0.36 (0.71) & 0.38 (0.69) & 0.25 (0.79) & 0.45 (0.7) &
% 0.65 (0.83) & 0.78 (0.88) & 0.68 (0.92) & 0.52 (0.68) &

% Llama-3.3-70B
% 0.74 (0.88) & 0.8 (0.94) & 0.8 (0.95) & 0.6 (0.77) &
% 0.66 (0.83) & 0.65 (0.79) & 0.68 (0.86) & 0.65 (0.79) &
% 0.33 (0.65) & 0.4 (0.72) & 0.11 (0.49) & 0.44 (0.68) &
% 0.72 (0.85) & 0.68 (0.9) & 0.84 (0.92) & 0.57 (0.71) &

% DeepSeek-V3
% 0.83 (0.98) & 0.83 (0.96) & 0.87 (1.0) & 0.82 (0.96)\\
% 0.82 (0.95) & 0.89 (0.97) & 0.84 (0.95) & 0.75 (0.9)\\
% 0.4 (0.81) & 0.42 (0.81) & 0.19 (0.76) & 0.5 (0.79)\\
% 0.76 (0.98) & 0.83 (1.0) & 0.73 (0.96) & 0.76 (0.95)\\

\begin{table*}
    \small
    \centering    
    \resizebox{1.0\textwidth}{!}{
        \begin{tabular}{l|cccc|cccc|cccc}
        \toprule
        \rowcolor[gray]{0.9} 
        Model & \multicolumn{4}{c|}{Llama-3.1-405B} & \multicolumn{4}{c|}{Llama-3.3-70B} & \multicolumn{4}{c}{DeepSeek-V3} \\
        \toprule
        Metrics & Total & OA & RC & SS & Total & OA & RC & SS & Total & OA & RC & SS \\
        \midrule
        \textit{Ours}
        & \textbf{0.66 (0.82)} & \underline{0.74 (0.84)} & \textbf{0.71 (0.88)} & \underline{0.52 (0.68)}
        & \textbf{0.74 (0.88)} & \textbf{0.8 (0.94)} & \underline{0.8 (0.95)} & \underline{0.6 (0.77)}
        & \textbf{0.83 (0.98)} & \underline{0.83 (0.96)} & \textbf{0.87 (1.0)} & \textbf{0.82 (0.96)}\\
        
        \textit{w/o Event}
        & 0.6 (0.71) & 0.46 (0.58) & \underline{0.68 (0.77)} & \textbf{0.6 (0.72)}
        & 0.66 (0.83) & 0.65 (0.79) & 0.68 (0.86) & \textbf{0.65 (0.79)}
        & \underline{0.82 (0.95)} & \textbf{0.89 (0.97)} & \underline{0.84 (0.95)} & 0.75 (0.9)\\
        
        \textit{Object List}
        & 0.36 (0.71) & 0.38 (0.69) & 0.25 (0.79) & 0.45 (0.7)
        & 0.33 (0.65) & 0.4 (0.72) & 0.11 (0.49) & 0.44 (0.68)
        & 0.4 (0.81) & 0.42 (0.81) & 0.19 (0.76) & 0.5 (0.79)\\
        
        \textit{Scene Graph}
        & \underline{0.65 (0.83)} & \textbf{0.78 (0.88)} & \underline{0.68 (0.92)} & \underline{0.52 (0.68)}
        & \underline{0.72 (0.85)} & \underline{0.68 (0.9)} & \textbf{0.84 (0.92)} & 0.57 (0.71)
        & 0.76 (0.98) & \underline{0.83 (1.0)} & 0.73 (0.96) & \underline{0.76 (0.95)}\\
        \bottomrule
        \end{tabular}
    }
    \caption{Additional benchmark result for test cases with \textit{object arrangement reasoning (OA)}, \textit{regional context reasoning (RC)}, and \textit{scene state reasoning (SS)} tags.}
    \label{tab:supp:benchmark_supp}
    \vspace{-1em}
\end{table*}

%% file: tab/benchmark_vlm.tex
% Llama-3.1-405B
% 0.66 (0.82) & 0.74 (0.84) & 0.71 (0.88) & 0.52 (0.68) &
% 0.6 (0.71) & 0.46 (0.58) & 0.68 (0.77) & 0.6 (0.72) &
% 0.36 (0.71) & 0.38 (0.69) & 0.25 (0.79) & 0.45 (0.7) &
% 0.65 (0.83) & 0.78 (0.88) & 0.68 (0.92) & 0.52 (0.68) &

% Llama-3.3-70B
% 0.74 (0.88) & 0.8 (0.94) & 0.8 (0.95) & 0.6 (0.77) &
% 0.66 (0.83) & 0.65 (0.79) & 0.68 (0.86) & 0.65 (0.79) &
% 0.33 (0.65) & 0.4 (0.72) & 0.11 (0.49) & 0.44 (0.68) &
% 0.72 (0.85) & 0.68 (0.9) & 0.84 (0.92) & 0.57 (0.71) &

% DeepSeek-V3
% 0.83 (0.98) & 0.83 (0.96) & 0.87 (1.0) & 0.82 (0.96)\\
% 0.82 (0.95) & 0.89 (0.97) & 0.84 (0.95) & 0.75 (0.9)\\
% 0.4 (0.81) & 0.42 (0.81) & 0.19 (0.76) & 0.5 (0.79)\\
% 0.76 (0.98) & 0.83 (1.0) & 0.73 (0.96) & 0.76 (0.95)\\

\begin{table*}
    \small
    \centering    
    \resizebox{0.9\textwidth}{!}{
        \begin{tabular}{l|cccc|cccc}
        \toprule
        \rowcolor[gray]{0.9} 
        Model & \multicolumn{4}{c|}{GPT-4o} & \multicolumn{4}{c|}{GPT-4o mini}\\
        \toprule
        Metrics & Total & OA & RC & SS & Total & OA & RC & SS\\
        \midrule
        \textit{Ours}
        & \textbf{0.9 (0.98)} & \textbf{0.93 (0.99)} & \textbf{0.9 (0.98)} & \textbf{0.92 (0.98)}
        & \textbf{0.74 (0.96)} & \textbf{0.72 (0.95)} & \textbf{0.72 (0.97)} & \textbf{0.78 (0.93)}\\
        
        \textit{Vision-based Description}
        & \underline{0.68 (0.97)} & 0.66 (0.98) & \underline{0.63 (0.96)} & \underline{0.76 (0.95)}
        & \underline{0.48 (0.92)} & {0.41 (0.93)} & \underline{0.47 (0.91)} & \underline{0.53 (0.86)}\\
        
        \textit{Vision-based Planning}
        & 0.67 (0.91) & \underline{0.69 (0.92)} & 0.52 (0.87) & 0.65 (0.86)
        & 0.38 (0.86) & \underline{0.43 (0.86)} & 0.15 (0.81) & 0.52 (0.84)\\
        \bottomrule
        \end{tabular}
    }
    \caption{Additional benchmark results for planning methods using vision-language models.}
    \label{tab:supp:benchmark_vlm}
    \vspace{-1em}
\end{table*}